\documentclass[sn-basic,Numbered]{sn-jnl}

\usepackage{enumerate}
\usepackage{amsmath}
\usepackage{indentfirst}
\usepackage{graphicx}%
\usepackage{multirow}%
\usepackage{amssymb}%
\usepackage{amsfonts}
\usepackage{amsthm}%
\usepackage{mathrsfs}%
\usepackage[title]{appendix}%
\usepackage{xcolor}%
\usepackage{textcomp}%
\usepackage{manyfoot}%
\usepackage{booktabs}%
\usepackage{algorithm}%
\usepackage{algorithmicx}%
\usepackage{algpseudocode}%
\usepackage{listings}%

\theoremstyle{thmstyleone}%
\theoremstyle{thmstyletwo}%

\theoremstyle{thmstylethree}%

\begin{document}

\title[Article Title]{From Bandits Model to Deep Deterministic Policy Gradient, Reinforcement Learning with Contextual Information}


\author*[1]{\fnm{Zhendong} \sur{Shi}}\email{shizd20@mails.tsinghua.edu.cn}

\author*[1]{\fnm{Ercan} \sur{Kuruoğlu}}\email{kuruoglu@sz.tsinghua.edu.cn}
\equalcont{These authors contributed equally to this work.}

\author[1]{\fnm{Xiaoli} \sur{Wei}}\email{xiaoli\_wei@sz.tsinghua.edu.cn}
\equalcont{These authors contributed equally to this work.}

\affil[1]{Tsinghua-Berkeley Shenzhen Institute, Tsinghua University, Shenzhen, China}


\abstract{The problem of how to take the right actions to make profits in sequential process continues to be difficult due to the quick dynamics and a significant amount of uncertainty in many application scenarios. In such  complicated environments, reinforcement learning (RL), a reward-oriented strategy for optimum control, has emerged as a potential technique to address this strategic decision-making issue. 
However, reinforcement learning also has some shortcomings that make it unsuitable for solving many financial problems, excessive resource consumption, and inability to quickly obtain optimal solutions, making it unsuitable for quantitative trading markets.
In this study, we use two methods to overcome the issue with contextual information: contextual Thompson sampling and reinforcement learning under supervision which can accelerate the iterations in search of the best answer. In order to investigate strategic trading in quantitative markets, we merged the earlier financial trading strategy known as constant proportion portfolio insurance (CPPI) into deep deterministic policy gradient (DDPG). 
The experimental results show that both methods can accelerate the progress of reinforcement learning to obtain the optimal solution.}




\maketitle

\section{Introduction}\label{sec1}

Compared with traditional trading methods, quantitative trading is widely known for its features of high-frequency, algorithmic, and automated trading, which is difficult to achieve by human beings in complex and dynamic stock market~\cite{bib19, bib7}.
In the quantitative market, massive noisy signals from stochastic trading behaviors and all kinds of unforeseeable social events make it difficult to predict the market state~\cite{bib10, bib15}.  Human traders can easily be affected by these events which would make the irrational decisions of trading nearly inevitable~\cite{bib16, 
bib8}. 
Therefore, different financial individuals and institutes from different research fields have started to explore more effective ways for handling these problems.

Dubey and Pentland \cite{bib25} proposed an algorithm for the MAB problem based on the symmetric $\alpha$-stable distribution \cite{samorodnitsky1997stable}. The approach demonstrated success through accurate assumptions and a normalized iterative process. The $\alpha$-stable distribution is a family of distributions characterized by heavy tails.

Motivated by the presence of asymmetric characteristics in various real life data \cite{kuruoglu03} and the success in reinforcement learning and other directions due to the introduction of asymmetry \cite{baisero2021unbiased}, in previous work, Shi et al. \cite{bib24} propose a statistic model, for which the reward distribution is both heavy-tailed and asymmetric, named asymmetric alpha-Thompson sampling algorithm. With the theorems for heavy-tailed distribution \cite{bib27}, variance analysis \cite{chen2016variance} and estimation methods for both symmetric and skewed $\alpha$-stable distribution \cite{bib26}, Shi et al. \cite{bib24} analyse the performances of symmetric $\alpha$-Thompson sampling algorithm and symmetric $\alpha$-Thompson sampling algorithm by Bayesian regret bound. In the general sequential decision making algorithm, asymmetric information\cite{gupta2014dynamic} is regraded as a single parameter derived from interdependence, common knowledge, higher order beliefs and so on. Our algorithm differs from other algorithms in that it assumes that the reward follows an asymmetric distribution, while others changing the their structures or using expert supervision to handle asymmetric information. Therefore, the range of our application is wider and more flexible.

Over the past years, with the development of artificial intelligence techniques, reinforcement learning (RL) has emerged as an efficient method for making decisions in dynamic environments with uncertainties~\cite{bib2}. 
The principle behind RL is the Markov decision process (MDP).
Through interacting with the environment, the RL agent, i.e. the decision maker, will iteratively update its strategy according to the rewards, which can be treated as guidance toward the expected target and the goal of the RL agent is hence to maximize the total reward~\cite{bib18}. 
Following the MDP, researchers from financial fields have tried to build their own specifically designed RL architecture to cope with different financial problems.
A deep RL method combined with knowledge distillation was proposed to improve the training reliability in the trading of currency pairs~\cite{bib20}.
To investigate the stock portfolio selection problem, a hypergraph-based RL method was designed to learn the policy function of generating appropriate trading actions~\cite{bib12}.
Besides, a policy-based RL framework for stock portfolio management was introduced and its performance was also compared with other trading strategies~\cite{bib21}.

The advantage of reinforcement learning in the financial field lies in its ability to cope with very complex market environments and uncertainties, adapt to constantly changing market conditions, and improve trading efficiency and profitability through continuous learning and adjustment of strategies. However, reinforcement learning also has limitations, such as requiring a large amount of computing resources and data, requiring good model and algorithm design, as well as stable data sources. In addition, due to the complexity of the financial market itself, reinforcement learning also faces many challenges in practice, such as overfitting, data sparsity and other problems, which need to be comprehensively considered and optimized in combination with practical application scenarios.

In this study, we use two methods to overcome the issue with contextual information: contextual Thompson sampling and a specific method to speed up reinforcement learning iterations in search of the best answer. In order to investigate strategic trading in quantitative markets, we merged the earlier financial trading strategy known as constant proportion portfolio insurance (CPPI) into deep deterministic policy gradient (DDPG) for investigating strategic trading in quantitative markets, respectively for studying how this novel architecture will behave in quantitative markets.

The Bandits model is a simplified reinforcement learning algorithm that has been widely applied in the financial field. By adjusting the distribution function to be heavy-tailed distribution and adding contextual information, we can make the bandits model more suitable for dealing with specific financial problems. Proportion portfolio insurance is a specific financial strategy designed for individuals with different risk preferences.

The rest of this work is organized as follows. Section 2 introduces the background knowledge, in which we discuss the similarities and differences between the contextual bandits model and reinforcement learning. The algorithms represent how contextual bandits work and how the DDPG specifically combined with CPPI strategy are shown in Section 3. Further, the numerical experiment and results are implemented and analyzed in Section 4. Finally, the conclusions are drawn in Section 5.

\section{Background Knowledge}\label{sec2}

In this section, we map the evolution from the bandits model to reinforcement learning and the similarities and differences between various algorithms. 

\begin{figure*}[ht]
    \centering
    \includegraphics[width=.78\linewidth]{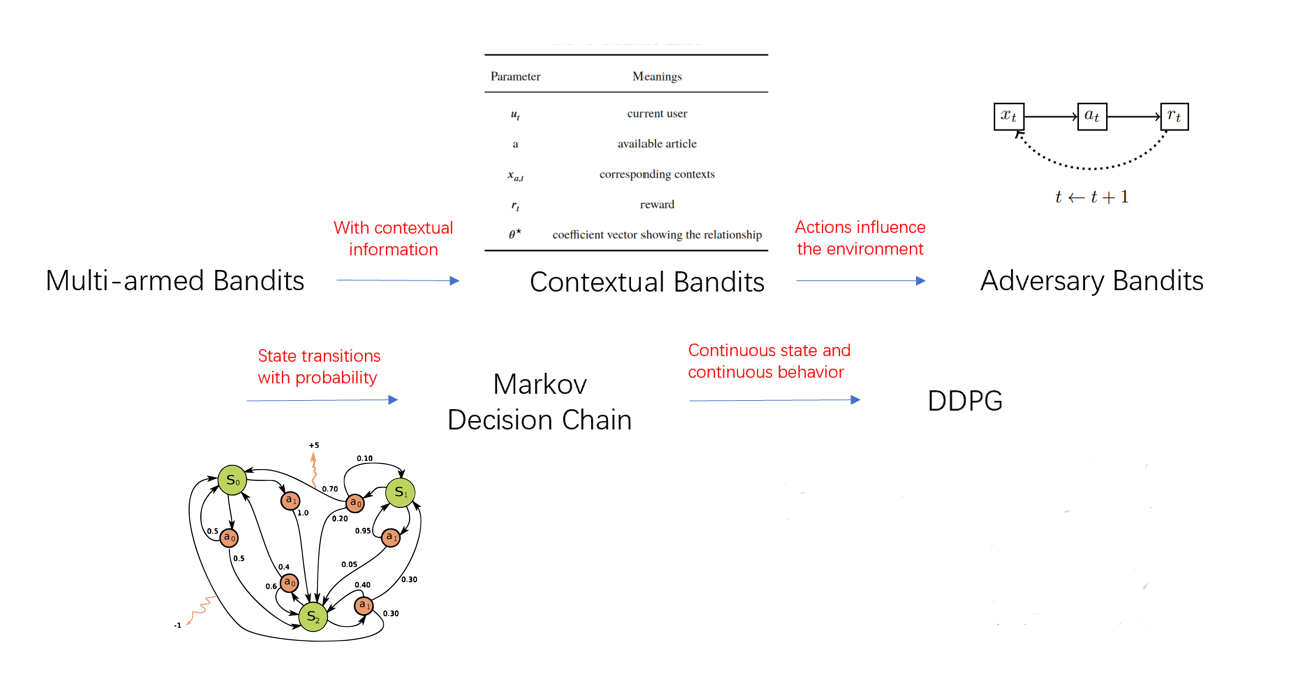}
    \caption{From bandits model to reinforcement learning}
    \label{fig:theo}
\end{figure*}

\subsection{Multi-Armed Bandit Problem}

Assume that an agent has a choice of multiple slot machines from which to choose for each round one's draw and recording of the payouts. If no two slot machines are precisely alike, we can gather some statistical data from each machine after several spins of the operations, and then choose the one that offers the largest projected payout.

The process of learning is indexed by $t$ $\in$ $[T]$. The entire number of rounds, denoted by the symbol $T$, is known beforehand. The agent selects an arm from $a_t$ $\in$ $[N]$, and in each round of $t$ $\in$ $[T]$, $r_{a_t}(t)$ is observed from that arm. Rewards are distributed individually for each arm $n$ $\in$ $[N]$ from a distribution $D_n$ with a mean $\mu_n$ = $E_{D_n} [r]$. The optimal arm(s) is(are) denoted as $n*$, and the associated arm(s) is(are) denoted as the biggest expected reward, $\mu_{\star}$ = $max_n \in [N]$ $\mu_n$.

The regret $R(T)$, which measures the discrepancy between the ideal total reward an agent can obtain and the total reward they actually receive, is used to measure performance.
        \begin{equation}
			\begin{aligned}
				R(T) =  \mu^{\star}T - \sum_{t=0}^{T} \mu_{a_t}.
			\end{aligned}
			\label{f1}
		\end{equation}
		

%
%

\subsection{Thompson Sampling Algorithm for Multi-Armed Bandit Problem}

A variety of exploration algorithms have been proposed, including $\epsilon$-greedy algorithm, UCB algorithm and Thompson sampling. $\epsilon$-greedy algorithm\cite{bib11} makes use of both exploitations to take advantage of prior knowledge and exploration in order to search for new solutions, whereas the UCB algorithm\cite{bib5} simply draws the arm that has the largest empirical estimate of reward up to that point plus some term that is inversely related to the number of times the arm was played.

Under the assumption that for each arm $n \in [N]$, the reward distribution is $D_n$ parametrized by $\theta_n \in \Theta$ ($\mu_n$ may not be an appropriate parameter) and that the parameter has a prior probability distribution p ($\theta_n$). The reward distribution is defined as follows: Thompson's sampling algorithm updates the prior distribution of $\theta_n$ as a function of the observed reward of $n$, and then chooses the arm according to the posterior probability derived from the reward under the arm $n$.

Through Bayes rule, 
\begin{equation}
	\begin{aligned}
	p(\theta|x) = \frac{p (x|\theta)p (\theta)}{p (x)} = \frac{p (x|\theta)p ( \theta)}{\int p (x|\theta^{'})p ( \theta^{'}) }d \theta^{'}.
	\end{aligned}
\label{f1221}
\end{equation}
where $\theta$ is the parameter and x is the observation. $p (\theta|x)$ is the posterior distribution, $p (x|\theta)$ is likelihood function, $p (\theta)$ is the prior distribution.

In each round, $t \in [T]$, the agent draws the $\hat{n}(t)$ parameter for each arm $n$ $\in$ $[N]$ according the posterior distribution of the parameters given the prior rewards up to time $t$, ${\pmb r_n} (t - 1)$ = $\{ r_n^{ (1)}, r_n^{ (2)}, \cdots, r_n^{ (k_n (t-1))}\}$, where $k_n (t)$ is how many times the arm $n$ has been pulled up at time $t$: 

\begin{equation}
	\begin{aligned}
		\hat{\theta}_n(t) \sim p(\theta_n|{\pmb r}_n(t-1)) \propto p({\pmb r}_n(t-1) | \theta_n) p(\theta_n).
	\end{aligned}
\label{f3}
\end{equation}
Through the parameters $\hat{\theta}_n$ (t) drawn from each arm, the agent chooses the arm $a_t$ with the highest mean return from the posterior distribution, receives the return $r_{a_t}$.
\begin{equation}
	\begin{aligned}
	a_t = \mathop{\arg\max}\limits_{n \in [N]} \mu_n (\theta_n(t))
\end{aligned}
\label{f4}
\end{equation}
		
In order to make a comparison with the symmetric case, we will use the Bayesian Regret \cite{bib17} for the performance measure.
The estimated regret with respect to the priors is Bayesian Regret (BR).
Denoting the parameters on the set of arms as $\overline{\theta} = \{\theta_1 , ..., \theta_N\}$ and their corresponding product distribution as $ \overline{D} = \prod_i D_i$, the Bayesian Regret is expressed in the following way. 
		\begin{equation}
			\begin{aligned}
				BR(T,\pi) = E_{\overline{\theta} \sim \overline{D}} [R(T)]
			\end{aligned}
			\label{f5}
		\end{equation}

\subsection{Alpha-Stable Distribution}

In many application scenarios, binary distribution and Normal distribution cannot accurately show the characteristics of the data set. The $\alpha$-stable distribution is a type of probability distribution that has a wide range of applications in various areas such as finance and signal processing. In finance, the $\alpha$-stable distribution is used to model fluctuations in asset prices and returns. It is particularly useful in modeling extreme events, such as stock market crashes or sudden changes in currency exchange rates. In signal processing, the $\alpha$-stable distribution is used for modeling noise in communication systems. It helps to understand how different types of interference can affect the quality of signal transmission.

The alpha-stable distribution is an important non-Gaussian distribution that is often used to model both impulsive and skewed data. It has a non-analytic density and therefore, usually is described with the characteristic function. We say a random variable X is $S_\alpha (\beta,\delta,\sigma)$ if X has characteristic function: 
\begin{equation}
\begin{aligned}
 \mathbb{E}[e^{iuX}]  = &\exp (-\sigma^{\alpha} \left|u\right|^{\alpha}(1+i \beta {\rm sign}(u) \\ & (\left|\sigma u\right|^{1-\alpha}-1)) + i u \delta )
\label{f6}
\end{aligned}
\end{equation}
$\alpha$ is used to indicate the impulsiveness of the distribution, parameter $\beta$ corresponds to the skewness, $\gamma$ is the scale parameter and $\mu$ is the mean, which is closely related with the location parameter delta via $\mu = \delta-\beta \sigma \tan(\pi \alpha/2)$. The shape parameter $\alpha$ must be in the interval $(0,2]$. When $\alpha>1$, the mean of the distribution exists and is equal to $\mu$, so this paper only concentrates on the interval $\alpha$ $\in$ (1,2) since the stable distribution will degenerate to Gaussian distribution when $\alpha$ is equal to 2.
\begin{figure}[h]
  \centering
  \includegraphics[width=0.5\linewidth]{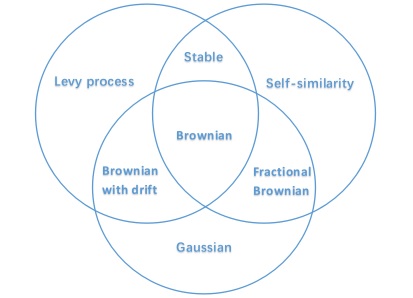}
  \caption{Stable distribution}
  \label{fig1}
\end{figure}

If $E[X^2] = \infty$, then X or its distribution has a heavy tail. X or its distribution has a heavy tail of order $\alpha$, where $\alpha \in (1,2)$, if $\lim_{t \rightarrow \infty} (|t|^{\alpha})P(|X| > t) = C < \infty$, $E|X|^p < \infty \iff p < \alpha$. When large and unexpected jumps occur between several relatively small observations, it is reasonable to suspect a heavy tail distribution. Stable distributions are as fundamental for heavy tail models as Gaussian distributions are for distributions with a finite second order moment. 

\subsection{Thompson Sampling Algorithm with Alpha-Stable Distribution}

The Thompson Sampling (TS) method received attention for its strong empirical evidence. This has led to more research on the algorithm’s theoretical analysis from a Bayesian standpoint. Specifically, Russo\cite{bib22} demonstrated the optimality of TS for Gaussian reward distributions. Korda\cite{bib23} later expanded on these findings to include a wider range of exponential family reward distributions.

The family of symmetric $\alpha$-stable distributions, known for their extremely heavy tails, were studied by Dubey and Pentland\cite{bib25} in Thompson Sampling. The study obtained the first polynomial regret bounds, independent of the problem, for Thompson Sampling when using symmetric $\alpha$-stable densities.

On the basis of Dubey and Pentland's work, Shi et al. \cite{bib24} extended the algorithm to $\alpha$-stable distributions through Gibbs Sampler and updating formula for $\alpha, \beta, \sigma, \delta$. Compared with symmetric $\alpha$-Thompson algorithm, asymmetric $\alpha$-Thompson algorithm can not only cover the asymmetry in data (which is very common in social data), but also greatly improve the accuracy of reward distribution assumptions by iterating on the four parameters of $\alpha$ stable distribution.

With the theorems for heavy-tailed distribution\cite{bib27} and estimation methods for both symmetric and skewed $\alpha$-stable distribution\cite{bib26}, Shi et al. \cite{bib24} analyse the performances of symmetric $\alpha$-Thompson sampling algorithm and symmetric $\alpha$-Thompson sampling algorithm by Bayesian regret bound. 

The subsequent algorithms in this paper are based on asymmetric $\alpha$-Thompson algorithm and are extensions of this algorithm.

\subsection{Contextual Information}

Contextual data refers to information that offers a perspective on an event, person, or thing by revealing how different pieces of data interrelate, resulting in a more comprehensive understanding of the subject. Such pertinent facts can be utilized to analyze behavior patterns, optimizing user experience.

For instance, companies may analyze sales data and incorporate information about traffic or weather conditions to gain a deeper understanding of the factors that impact their sales. In the realm of big data, information without context often lacks practical value. By incorporating contextual information, organizations can make more informed and accurate decisions at a higher level.

As for the value in the algorithm, we use context data to show the action mode when it is too complex to be covered by a single set of data. Many factors and their relationships need to be considered for description of the action mode of these things, which is the dimension upgrading processing of simple data.

\subsection{Thompson Sampling for Contextual Bandits}

The Contextual Multi-Armed Bandit (CMAB), also known as the Contextual Bandit, is a useful variation of the multi-armed bandit problem. This scenario assumes that the agent observes an N-dimensional context or feature vector prior to selecting an arm at each iteration. In other words, the goal of the learner in CMAB is not only to maximize reward but also to learn the relationship between the feature vectors and the rewards, in order to make better decisions in the future. This is especially useful in real world applications such as personalized recommendations, where the context may be the user's past behaviour, interests and demographics.

Some recent studies have also considered the non-linear relationship between the actions’ expected reward and their contexts, such as Neural Bandit, LINUCB \cite{bib6}, and Contextual Thompson Sampling \cite{bib1}. These algorithms typically assume a linear relationship between an action’s expected reward and its context.

CMAB algorithms have found application in a variety of real-world problem settings, including healthcare, computer network routing, and finance, among others. CMAB approaches are also useful in computer science and machine learning for hyper-parameter tuning and algorithmic choices in supervised learning and reinforcement learning. 

In multi-armed bandit (MAB) problems, a random reward $r_i(t)$ with an unknown mean $\theta_i(t)$ is assumed. In the contextual MAB problem, it is assumed that $\theta_i(t) = \theta_t b_i(t)$, where $\theta_i(\cdot)$ is an arbitrary function and $b_i(t)$ is a context vector. Specifically, it's a Linear contextual MAB problems if we assume that $\theta_t b_i(t)$ is linear in $b_i(t)$:

\begin{equation}
	\begin{aligned}
		\theta_t b_i(t) = b_i(t)^T \mu , i =1, ..., N
	\end{aligned}
    \label{f34331}
    \end{equation}

The key advantages of linear contextual bandits is their simplicity and efficiency. They can handle large datasets with many features and do not require complicated optimization techniques. Additionally, they can operate with incomplete information, which is useful when data is missing or incomplete.

\subsection{Non-Linear Contextual Bandits}

In contextual bandits, a non-linear function is used to model the relationship between the context and the expected reward, rather than a linear function. However, the function is only partially specified, which allows for flexibility and adaptation to different contexts. The parameters of the function are learned from data through a process of trial and error, using a limited subset of the available context information\cite{bib3}.
\begin{equation}
	\begin{aligned}
		\theta_t b_i(t) = b_i(t)^T \mu +v(t), i =1, ..., N 
	\end{aligned}
    \label{f34331}
    \end{equation}
If $v(t)=0$, the contextual multi-armed bandit (MAB) problem  reduces to a linear contextual MAB. When $v(t)$ is not zero but unrelated to the action, it is referred to as a semi-parametric MAB problem.

If $v(t)$ depends on the action, then the reward distribution becomes completely non-parametric, and the problem is defined as adversarial. In adversarial contextual bandits\cite{bib3}, an “adversary” is introduced to the environment and tries to actively manipulate the reward signals received by the agent. Under changing conditions, the agent must learn a policy that is robust to the adversary's manipulations. This is often achieved through a process of exploration and adaptation, where the agent continually adjusts its behavior to minimize the impact of the adversary’s actions.

Contextual adversarial bandits are simplified versions of complete reinforcement learning problems with n arms. The adversarial contextual bandits\cite{bib9} avoid the complexity of full RL by teaching the agent to act in only one situation with n different possible actions or options. 

\subsection{Deep Deterministic Policy Gradient(DDPG)}

A sequential decision-making problem in the scenario can be described as a stochastic game, which can be defined by a set of key elements $\mathbb{S},\mathbb{A}$, $P,R$ ,$\gamma \rangle$. At time $t$, under a shared state $S_t\in\mathbb{S}$, each agent takes its action $a\in\mathbb{A}$ simultaneously. The action $\mathbf{a}$ leads the environment changes according to the dynamics $P:\mathbb{S}\times\pmb{\mathbb{A}}\rightarrow \Delta(\mathbb{S})$. After that each agent receives its individual reward $r^i$ according to its reward function $R^i:\mathbb{S}\times\pmb{\mathbb{A}}\times\mathbb{S}\rightarrow \mathbb{R}$. $\gamma$ is the discount factor that represents the value of time.

In our setting, we have one agent trading in quantitative markets using different strategies. Its goal is to maximize individual return while keeping a certain degree of diversity among the portfolios managed by each agent since we want to allocate risks. The state space shared by our agents is the raw historical closing prices of all stocks. At each step, our agent will output an action of choosing several stocks with the amount of operation (i.e., buy, sell or hold).
\begin{figure}[ht]
\centering
\includegraphics[width=\linewidth]{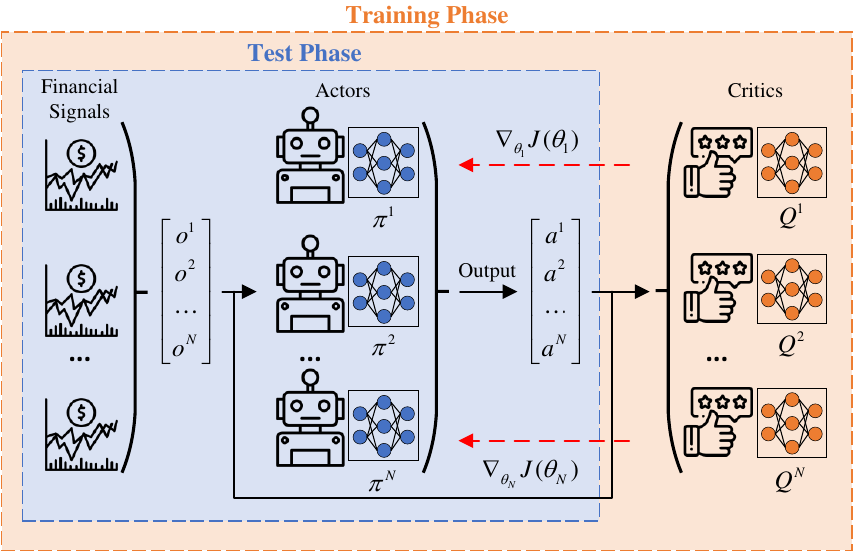}
\caption{A schematic of DDPG in the quantitative market environment.}
\label{MADDPG_frame}
\end{figure}

We adopt the DDPG~\cite{bib12} to train our agents. This approach is specifically established for the implementation in the scenario of quantitative trading, presented as Fig.~\ref{MADDPG_frame}. DDPG is a  version of actor-critic method considering a continuous action set with a deterministic policy, where each agent has an actor network $\pi_\theta$ parameterized by $\theta$ and a critic network $Q_\phi$ parameterized by $\phi$. Each agent learns its optimal policy by updating the parameters of its policy networks to directly maximize the objective function, i.e., the cumulative discounted return, $J(\theta_i)=\mathbb{E}_{s\sim P,a\sim\pi_\theta}[\sum_{t\geq 0}\gamma^t R_t^i]$ and the direction to take steps by agent $i$ can be presented as the gradient of the cumulative discounted return, shown as follow equations:
\begin{equation}\label{actor loss}
\nabla_{\theta_i}J(\theta_i)=\mathbb{E}_{s\sim\mathcal{D}}\bigg[\nabla_{\theta_i}\log \pi_{\theta_i}(a_i|s)\cdot\\
    \nabla_{a_i}Q_{\phi_i}(s,a)|_{a=\pi_{\theta}(s)}\bigg]
\end{equation}
where $\mathcal{D}$ is the experience replay buffer containing tuples $(s,s',a,r)$ that are stored throughout training. The centralized critic networks are updated by approximating the true action-value function using temporal-difference learning,
\begin{equation}
    \mathcal{L}(\phi_i)=\mathbb{E}_{s,a,r,s'}\bigg[(Q_{\phi_i}(s,a)-y)^2\bigg], \quad
    y=r_i+\gamma Q_{\phi'_{i}}(s',a')|_{a'_{j}=\pi_{\theta'_{i}}(s)}
\end{equation}
where $Q_{\phi'_{i}}$ and $\pi_{\theta'_{i}}$ are target networks with delayed parameters $\phi'_{i}$ and $\theta'_{i}$ like in deep Q-network method~\cite{bib14}. The purpose of introducing target networks is to ease the moving target problem in deep reinforcement learning and stabilize the off-policy learning procedure.

The core idea of DDPG is to centralize training while execution in a decentralized manner. The centralized critic networks $Q$ utilize past action. When execution, only the actor network is used to generate policy. This technique serves as a cure to the non-stationarity problem in MARL.   

\subsection{Strategies of Constant Proportion Portfolio Insurance(CPPI)} %

In many reinforcement learning frameworks, we have access to data from the system operated by its predecessor controller, but we do not have access to a precise simulator of the system. For this reason, we want the agent to learn as much as possible from the demonstration data prior to running on the actual system. In the pre-training phase, the goal is to learn how to imitate the demonstrator with a value function that satisfies Bellman's equation. 

\begin{figure}[ht]
\centering
\includegraphics[width=\linewidth]{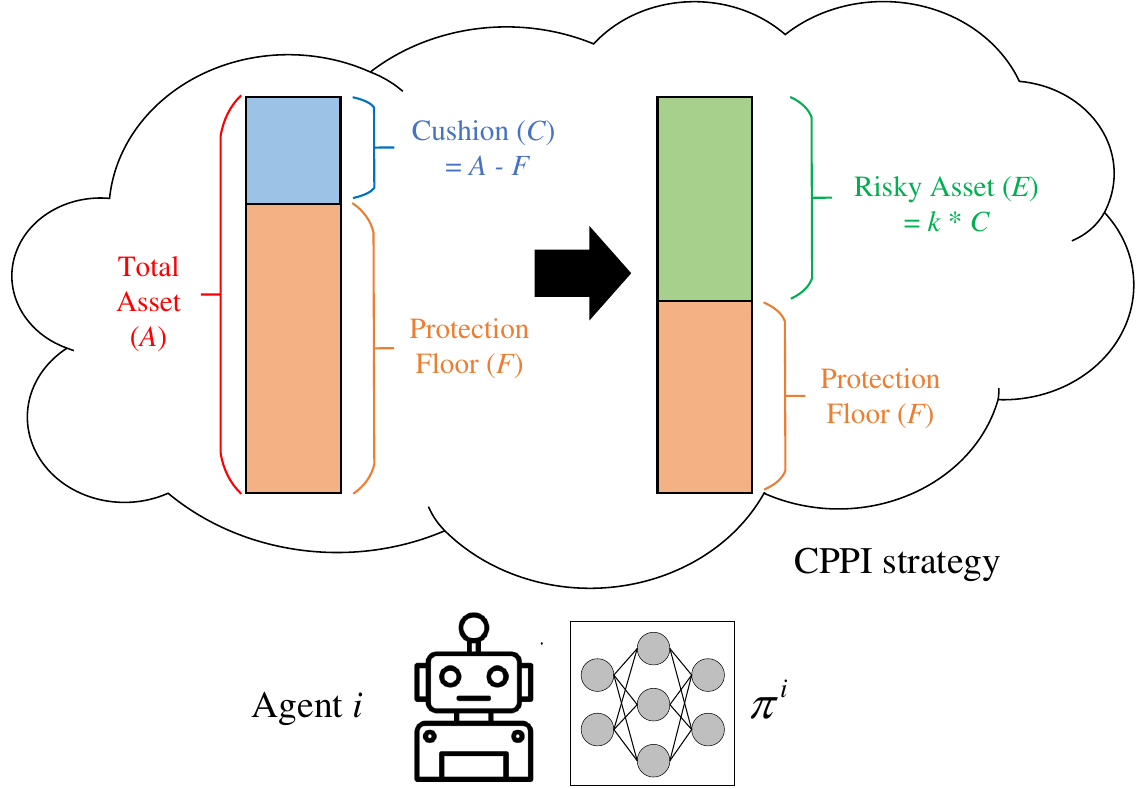}
\caption{The principle of CPPI strategy for agent $i$.}
\label{CPPI}
\end{figure}

CPPI is a type of portfolio insurance in which the investor sets a floor based on their asset, then structures asset allocation around the trading decision~\cite{bib4}. 
As shown in Fig.~\ref{CPPI}, the total asset $A$ is separated into two parts, the protection floor $F$ and the cushion $C$, in which the floor $F$ is the minimum guarantee used for protecting the basis of the total asset and the multiple cushions $k*C$ is supposed to be used as the risky asset $E$,

\begin{equation}
    E = k * C = k * (A - F),
\end{equation}
where the risk factor $k$ indicates the measurement of the risk and a higher value denotes a more aggressive trading strategy.

For pre-training to have any effect, supervised loss is crucial. As the demonstration data necessarily covers a narrow portion of the state space and does not take all possible actions, many state-actions have never been taken and have no data to ground them to realistic values. To obtain the new loss function, we add a large margin classifier loss. 
\begin{equation}
   J_{E}(Q) = max_{a \in A}[Q(s,a)+l(a_{E},a)]-Q(s,a_{E})
\end{equation}
where $a_{E}$ is the action the expert demonstrator took in state s and $l(a_{E},a)$ is a margin function that is 0 when $a = a_{E}$ and positive otherwise.  

The overall loss is used to combine losses to update the network:

\begin{equation}
  J(Q) = J_{DDPG}(Q) + \lambda J_{E}(Q)
\end{equation}
where $\lambda$ parameter shows the weighting between the losses.

\section{Algorithms}\label{sec3}

\subsection{Asymmetric Alpha-Thompson Sampling with Contextual Bandits}

\subsubsection{Parameter Setting}

Assume the mean of reward $\theta_t b_i (t)$ and $b_i (t)$ is linear: $\theta_t b_i (t)=[b_i (t)]^T \mu$, $i=1,...,N$ and $\mu$ is unknown. Assuming the expectation $ r_i (t) = [b_i (t)]^T \mu$ (linear component that does not change with time, depending on action ) $+ v(t)$ (nonparametric component over time, possibly depending on history information, but not action dependent)

The distribution of v(t) (disturbance term) is assumed unknown, if v(t) = 0, linear Contextual Multi-Armed Bandit; otherwise, if v(t) also depends on the action, Adversarial Contextual Multi-Armed Bandits. 

These models are part of generalized linear bandits that have binary rewards, represented as $A$ for the set of arms. The algorithm is an extension of Thompson Sampling that assumes Bernoulli reward distributions. It models the expected reward at each time step using a logistic function, denoted as $\mu$, that depends on the context $x$ and a parameter vector $\theta$ in $\mathbb{R}^D$. Specifically, the probability of receiving a reward of 1 for selecting arm $a$ at time $t$ is represented by $p_{a,t} = \mu(\boldsymbol{\theta}a^T x_t)$.

Then we apply an extended version of Bayesian Thompson Sampling. This method maintains a posterior distribution over arm parameters, which is updated based on observed rewards. At each time step, an arm with the highest probability of achieving the highest expected reward is selected.

This algorithm balances exploration and exploitation, and can achieve near-optimal cumulative regret. It has shown good performance in a variety of contexts, outperforming other popular algorithms such as UCB and epsilon-greedy. In Bayesian terminology, exploration can be seen as the process of updating the prior distribution based on new data or information. This corresponds to the Bayesian notion of updating beliefs based on evidence. Exploitation, on the other hand, can be seen as the process of using the current knowledge or belief (i.e., the prior) to make decisions that maximize the expected reward. This corresponds to the Bayesian notion of using the posterior distribution to make decisions that maximize the expected utility.

In a news recommendation system, the algorithm selects one article from a discrete set of articles $A$ in each trial at time $t=1,2,…$, and the reward is obtained when the user clicks on the recommended article (1 if the user clicks, 0 otherwise).

Considering contextual bandit problems for article recommendation, articles and users are characterized by contextual attributes, such as genre and popularity for articles, or age and gender for users. At each trial $t$, the learner observes the current user $u_t$, the set of available articles $A$, and their respective contexts $x_{t,a}$, which are d-dimensional summaries of both the user and article contexts.

The objective at each time step in contextual bandit problems for article recommendation is to recommend an article to the current user (i.e., select an arm from the set $A$), and the subsequent user action, whether they click or not, results in a binary reward of 1 or 0. The correlation between the reward and the contextual attributes is mediated by a coefficient vector $\boldsymbol{\theta}^*$, which reflects the user’s preferences concerning different article attributes.

This model can be used in a variety of scenarios, such as news recommendation, where the articles are characterized by their topics and authors and the users by their reading history and demographic information. The learner aims to balance exploration of unfamiliar articles with exploitation of previously successful recommendations to maximize cumulative rewards.
\begin{table}[ht!]
    \normalsize
    \caption{Parameter Defination}
    \label{tab:data}
    \centering
    \begin{tabular}{c c}
      \toprule
      Parameter & Meanings \\
      \midrule 
     $u_t$ &  current user \\
     a  & available article \\
     $x_{a,t}$ & corresponding contexts \\
     $r_t$ & reward \\
     $\theta^{\star}$ & coefficient vector showing the relationship\\
      \bottomrule
    \end{tabular}
\end{table}

The baseline tendency in user’s clicking behavior can change unexpectedly due to different users visiting at each time, and even for the same user, the clicking tendency can modify based on their mood or schedule, which cannot be captured as contextual information. Hence, the probability of the user clicking on an article is assumed to be linearly associated with the contextual information of the article and user, given this baseline tendency.

\subsubsection{Algorithm for Asymmetric Alpha-Thompson Sampling with Linear Contextual Bandits}

The main difference between contextual multi-armed bandits algorithm and non-contextual one is that the linear relationship leads to the uncertainty of $\theta$. For non-contextual bandits algorithms, each arm has a certain parameter $\theta$, while for contextual bandits algorithms, the parameters $\theta$ of arms for agents with different preferences are different.

\begin{algorithm}[H]
\caption{Contextual Thompson Sampling}
    \hspace*{0.01in} 
    Set $B = I_{d}, r = 0_{d},$ d is the dimension of context vector. 
	\begin{algorithmic}[1]
    \For{each iteration $t$ $\in$ $[T]$}
        \State Compute $\mu(t) = B^{-1} r.$
        \State Sample $\hat{\mu(t)}$ from Distribution $N(\mu(t),v^2 B^{-1})$ 
        \State Pull arm $a(t) = argmax_{i \in [N]} {\theta_i(t)}^{T} \mu(t) $ and get reward $r_{a(t)}(t)$.
        \For{each arms $n$ $\in$ [N]}
            \State Compute $\pi(t) = P(a(t)=i|F_{t-1})$
        \EndFor
        \State Update B and r:
        \State $B \longleftarrow B+ (\theta_{a(t)}(t)-\overline{\theta}(t))(\theta_{a(t)}(t)-\overline{\theta}(t))^{T}+ \sum_{k=1}^{N} \pi_i(t) (\theta_{k}(t)-\overline{\theta}(t))(\theta_{k}(t)-\overline{\theta}(t))^{T}$
        \State $r \longleftarrow r+ 2*(\theta_{a(t)}(t)-\overline{\theta}(t)) r_{a(t)}(t)$
    \EndFor

    \end{algorithmic}
\end{algorithm}

In general,  we assume that reward obeys normal distribution. At this time, according to the conjugation of Bayesian formula, we get the formula:
\begin{equation}
 \begin{split}
     \theta = (B^T B + I)^{-1} B^T \mu
 \end{split}
\end{equation}
As for asymmetric alpha-stable distribution, we need the new Bayesian inference formula.

The steps of the algorithm can be explained as follows:

\begin{enumerate}[a]
    \item  Estimate initial parameter for $\alpha$, $\beta$ and $\sigma$ and prior distribution $p(\delta) $ (Line 1)
 
    \item  In each round, use the sampled parameter vectors to calculate the expected reward for each arm (Line 3)

    \item  Select the arm with the highest expected reward (Line 4)
    
	\item   Decide whether accept the drawn theta from prior distribution through Metropolis–Hastings algorithm and the relationship between prior and posterior distribution (Line 9)
 
	\item   Get reward distirbution which is decided by new theta (Line 11),
  \item Update the posterior distribution for all parameters, in the linear condition, the change is shown by the update of B and r. (Line 13 - 16)
  \end{enumerate}

\begin{algorithm}[H]
\caption{Contextual Asymmetric $\alpha$-Thompson Sampling}
    \hspace*{0.01in} 
    Set $B = I_{d}, r = 0_{d},$ where d is the dimension of context vector. 
    Arms n $\in$ [N], priors $\alpha, \beta, \sigma$ for each arm, auxiliary variable y
	\begin{algorithmic}[1]
    
    \State estimate all $\alpha, \beta, \sigma$ by empirical characteristic function method and deduce prior distribution $p(\delta)$
    \For{each iteration $t$ $\in$ $[T]$}
        \State Compute $\mu(t) = B^{-1} r.$
        \State Pull arm $a(t) = argmax_{i \in [N]} {\theta_i(t)}^{T} \mu(t) $ and get reward $r_{a(t)}(t)$.
        \For{each arms $n$ $\in$ [N]}
            \State draw $\theta_n(t)$ from prior distribution
            \For{dimension j $\in$ [d]}
                \State Generate $u$ from a Uniform(0,1)
                \State If $u \textless  p (\hat{\theta}_n^{j}(t) | \alpha, \beta, \sigma, {\pmb r}_n^{j}(t)) p(\theta_n^{j}(t)|\hat{\delta}_n^{j}(t)) / (p(\theta_n^{j}(t) | \alpha, \beta, \sigma, {\pmb r}_n^{j}(t)) $  $ p(\hat{\theta}_n^{j}(t)|\theta_n^{j}(t)))$ then $\theta_n^{j}(t+1) = \hat{\theta}_n^{j}(t)$; otherwise, $\theta_n^{j}(t+1) = \theta_n^{j}(t)$
            \EndFor
            \State Compute $P_{n}(t) = \int_{\theta_{a(t)}}^{+\infty} p(z|\theta_{n})dz$
        \EndFor
        \State Update B and r:
        \State $B \longleftarrow B+ (\theta_{a(t)}(t)-\overline{\theta}(t))(\theta_{a(t)}(t)-\overline{\theta}(t))^{T}+ \sum_{k=1}^{N} P_{k}(t)/(\sum_{k=1}^{N} P_{k}(t))(\theta_{k}(t)-\overline{\theta}(t))(\theta_{k}(t)-\overline{\theta}(t))^{T}$
        \State $r \longleftarrow r+ 2*(\theta_{a(t)}(t)-\overline{\theta}(t)) r_{a(t)}(t)$
        \State Update posterior distribution $p(\delta_n(t+1))$ $p(\alpha_n(t+1))$ $p(\beta_n(t+1))$ $p(\sigma_n(t+1))$ by equations
        
    \EndFor

    \end{algorithmic}
\end{algorithm}

\subsubsection{Algorithm for Asymmetric Alpha-Thompson Sampling with Semi-Parametric Contextual Bandits}

The central concept behind conditioning is that the non-stationarity of the rewards does not vary significantly across arms. Therefore, centering the context around the mean for each arm does not alter the problem of selecting the arm with the highest expected reward. This enables the construction of an estimator for $\mu_j$ that is robust to the effect of $v_j (t)$, while simultaneously utilizing user affinity information through the creation of a graph.

In this model, conditioning is a method that accounts for the variation in rewards across different contexts and helps to reduce exploration time by exploiting the knowledge gained from exploring other similar contexts. By conditionally centering the context around the mean of each arm, we can focus on the differences between the arms and choose the one that is most likely to yield the highest reward.

The graph structure can be utilized to model user affinity information by connecting users with similar preferences, thereby facilitating the transfer of knowledge between users in the form of reward information. This approach can lead to faster learning and improved performance, as it allows the algorithm to exploit previously acquired knowledge rather than starting from scratch in each new context.

\begin{algorithm}[h]
\caption{Semi-Contextual Thompson Sampling}
    \hspace*{0.01in} 
    Fix $\lambda >0 $ $B_j(1) = \lambda l_{jj} I_{d}, y_j(1) = 0_{d},$ d is the dimension of context vector. 
	\begin{algorithmic}[1]
    \For{each iteration $t$ $\in$ $[T]$}
        \State Observe $j_t$
        \For{each iteration $j$ $\in$ $[N]$}
            \If{j $\neq j_t$} 
                \State UPDATE $B_j(t+1) \longleftarrow B_j(t)$, $\overline{\mu_j(t+1) } \longleftarrow\overline{\mu_j(t)}$, and $y_j(t+1) \longleftarrow y_j(t)$
            \Else
                \State $\hat{\mu_j(t)} \longleftarrow \overline{\mu_j(t)} - B_j(t)^{-1} \sum_{k \neq j} \lambda l_{jk} \overline{\mu_k(t)}$ 
                \State $\Gamma_j(t) \longleftarrow B_j(t) + \lambda^2 \sum_{k \neq j} l_{jk}^2 B_k(t)^{-1}$ 
                \State Sample $\mu_j(t) $ from $N(\hat{\mu_j(t)}, v_j^2 \Gamma_j(t)^{-1}$
                \State Pull arm $a(t) = argmax_{i \in [N]} {\theta_i(t)}^{T} \mu_j(t) $ and get reward $r_{a(t),j}(t)$.
                \State $\pi_i(t)  \longleftarrow P(a(t)=i|F_{t-1}) i \in [N] $
                \State $\overline{\theta(t)} \longleftarrow \sum_{i=1}^N \pi_i(t) \theta_i(t) $ and $X \longleftarrow theta_{a(t)}(t) - \overline{\theta(t)}$
                \State Update B, y and $\mu$:
                \State $B_j(t+1) \longleftarrow B_j(t) + (\theta_{a(t)}(t)-\overline{\theta}(t))(\theta_{a(t)}(t)-\overline{\theta}(t))^{T}+ \sum_{k=1}^{N} \pi_i(t) (\theta_{k}(t)-\overline{\theta}(t))(\theta_{k}(t)-\overline{\theta}(t))^{T}$
                \State $y_j(t+1) \longleftarrow y_j(t) + 2 X(t) r_{a(t),j}(t)$, and $\overline{\mu_j(t+1)} = B_j(t+1)^{-1} y_j(t+1)$
            \EndIf
        \EndFor
    \EndFor
    \end{algorithmic}
\end{algorithm}

The main difference with the Asymmetric $\alpha$-Thompson Sampling is shown by Line 7-8. It is important to note that the proposed estimator $\hat{\mu}_j(t)$ and the subsequent Thompson sampling step are both local, in the sense that they are only executed for user $j_t$ at each time step, and not for all users simultaneously. This approach is motivated by the fact that we typically lack updated information about other users at time $t$.

The idea of local updates is a natural consequence of the information available at each time step, as only the current user and their corresponding context are observed. By focusing on the local user, we can effectively utilize the available information and tailor the recommendations to the user’s preferences. Through the proposed estimator $\mu_j(t)$, a combination of single user based semi-parametric contextual and asymmetric alpha-stable assumption is derived.

The main difference with the Asymmetric $\alpha$-Thompson Sampling is shown by Line 8-9. With the assumption that the context around the mean for each arm does not alter the problem of selecting the arm with the highest expected reward, we have minimized the impact of terms $v(t)$ through new estimator $\mu_{j}(t)$.

\begin{algorithm}[H]
\caption{Semi-Contextual Asymmetric $\alpha$-Thompson Sampling}
\label{algo}
    \hspace*{0.01in} 
    Fix $\lambda >0 $ $B_j(1) = \lambda l_{jj} I_{d}, y_j(1) = 0_{d},$ d is the dimension of context vector.
    Arms n $\in$ [N], priors $\alpha, \beta, \sigma$ for each arm, auxiliary variable y
	\begin{algorithmic}[1]
    \State estimate all $\alpha, \beta, \sigma$ by empirical characteristic function method and deduce prior distribution $p(\delta)$
    \For{each iteration $t$ $\in$ $[T]$}
        \State Observe $j_t$
        \For{each iteration $j$ $\in$ $[N]$}
            \If{j $\neq j_t$} 
                \State Update $B_j(t+1) \longleftarrow B_j(t)$, $\overline{\mu_j(t+1) } \longleftarrow\overline{\mu_j(t)}$, and $y_j(t+1) \longleftarrow y_j(t)$
            \Else
                \State $\hat{\mu_j(t)} \longleftarrow \overline{\mu_j(t)} - B_j(t)^{-1} \sum_{k \neq j} \lambda l_{jk} \overline{\mu_k(t)}$ 
                \State Sample $\mu_j(t) $ from $N(\hat{\mu_j(t)}, v_j^2 \Gamma_j(t)^{-1}$
                \State Pull arm $a(t) = argmax_{i \in [N]} {\theta_i(t)}^{T} \mu_j(t) $ and get reward $r_{a(t),j}(t)$.
                \State draw $\theta_n(t)$ from prior distribution
                \State Generate $u$ from a Uniform(0,1)
                \State If $u \textless  p (\hat{\theta}_n^{j}(t) | \alpha, \beta, \sigma, {\pmb r}_n^{j}(t)) p(\theta_n^{j}(t)|\hat{\delta}_n^{j}(t)) / (p(\theta_n^{j}(t) | \alpha, \beta, \sigma, {\pmb r}_n^{j}(t)) $  $ p(\hat{\theta}_n^{j}(t)|\theta_n^{j}(t)))$ then $\theta_n^{j}(t+1) = \hat{\theta}_n^{j}(t)$; otherwise, $\theta_n^{j}(t+1) = \theta_n^{j}(t)$
            \State Compute $P_{n}(t) = \int_{\theta_{a(t)}}^{+\infty} p(z|\theta_{n})dz$
            \State Update B and r:
            \State $B_j(t+1) \longleftarrow B_j(t) + (\theta_{a(t)}(t)-\overline{\theta}(t))(\theta_{a(t)}(t)-\overline{\theta}(t))^{T}+  P_{j}(t)/(\sum_{k=1}^{N} P_{k}(t))(\theta_{k}(t)-\overline{\theta}(t))(\theta_{k}(t)-\overline{\theta}(t))^{T}$
            \State $y_j(t+1) \longleftarrow y_j(t) + 2 X(t) r_{a(t),j}(t)$, and $\overline{\mu_j(t+1)} = B_j(t+1)^{-1} y_j(t+1)$
            \State Update posterior distribution $p(\delta_n(t+1))$ $p(\alpha_n(t+1))$ $p(\beta_n(t+1))$ $p(\sigma_n(t+1))$ by equations
            \EndIf
       \EndFor
        
    \EndFor

    \end{algorithmic}
\end{algorithm}

\subsubsection{Algorithm for Asymmetric Alpha-Thompson Sampling with Adversarial Contextual Bandits}

Adversarial contextual bandits can be regarded as a simplified RL problem, in which an adversary modifies the reward function based on the actions taken by the learner. This can be seen as a form of exploration-exploitation trade-off, as the learner decides between exploring new actions and exploiting previously successful actions, while the adversary tries to prevent the learner from obtaining high rewards.

Despite its simplicity, the adversarial contextual bandit problem is highly relevant in practical applications, such as online advertisement, where the reward function is affected by user behavior and external factors, making it difficult to achieve optimal performance. 

In more complex environments, the dependence between the observed contextual information and the chosen action of the agent might not be independent, and can be modeled as a Markov decision process (MDP) in a reinforcement learning (RL) problem.

In an MDP, the agent observes a state $s_t$ at time step $t$ and then selects an action $a_t$ based on their policy. The environment responds to the action by transitioning to a new state $s_{t+1}$ and providing the agent with a reward $r_{t+1}$.

We consider a simple extension of our analysis to contextual episodic Markov decision process (MDP) with unknown but deterministic transitions, denoted by
\begin{equation}
 \begin{split}
     M = MDP(S,A,H,P,r)
 \end{split}
\end{equation}
The player interacts with this contextual episodic MDP as follows. In each episode $t = 1, . . . , T $, a context $x_t^1 \in S^1 \subset S$ is picked arbitrarily by an adversary.

\noindent The goal of MDP is to optimize the expected cumulative rewards:
\begin{equation}
 \begin{split}
    E \sum_{t=1}^T \sum_{h=1}^H r_t^h
 \end{split}
\end{equation}
It is known that the optimal policy can be derived from the Q function of the MDP. 
\begin{equation}
 \begin{split}
    Q^h (x^h,a^h) = E[r^h|x^h,a^h] + max_{a^{h+1}} Q^{h+1} (x^{h+1},a^{h+1})
 \end{split}
\end{equation}
For simplicity, we assume that $Q^{H+1} (·) = 0$. 

\noindent The regret of an MDP algorithm at each time step t is defined as:
\begin{equation}
 \begin{split}
    REGRET_t = max Q^1(x^1,a^1) - E \sum_{h=1}^H r_t^h
 \end{split}
\end{equation}

The flowchart of MDP-contextual bandits can be broken down into the following steps:

\begin{enumerate}[a]
    \item  The algorithm starts by estimating the parameters and initializing the state, action, and reward history (Line 1)
 
    \item  At each time step, the algorithm receives a context, calculates the expected reward for each arm and selects an action based on the current policy (Line 9)

    \item  After selecting an action, the algorithm receives a reward and a new context. The reward and context are used to update the state of the MDP (Line 10-12)
    
	\item  The algorithm updates the postrior distribution of parameters and value function $Q$ of the MDP based on the current state and the rewards received (Line 13-14)
 
	\item   The algorithm improves the policy based on the updated value function. (Line 15)

\end{enumerate}

The main difference with Semi-Contextual $\alpha$-Thompson Sampling is the Q function which makes future actions have an impact on current choices.

\begin{algorithm}[H]
\caption{MDP-Contextual Asymmetric $\alpha$-Thompson Sampling}
    \hspace*{0.01in} 
    Arms n $\in$ [N], priors $\alpha, \beta, \sigma$ for each arm, auxiliary variable y
	\begin{algorithmic}[1]
    \State estimate all $\alpha, \beta, \sigma$ by empirical characteristic function method and deduce prior distribution $p(\delta)$
    \For{each iteration $t$ $\in$ $[T]$}
        \State Observe $j_t \in S$
        \For{each iteration $j$ $\in$ $S$}
            \State Draw $\theta_j(t) ~ p(·|S_{j,t-1})$
            \If{j $\neq j_t$} 
                \State Update $B_j(t+1) \longleftarrow B_j(t)$, $\overline{\mu_j(t+1) } \longleftarrow\overline{\mu_j(t)}$, and $y_j(t+1) \longleftarrow y_j(t)$
            \Else
                \State refer to Algorithm \ref{algo}
            \State Update B and r:
            \State $B_j(t+1) \longleftarrow B_j(t) + (\theta_{a(t)}(t)-\overline{\theta}(t))(\theta_{a(t)}(t)-\overline{\theta}(t))^{T}+  P_{j}(t)/(\sum_{k=1}^{N} P_{k}(t))(\theta_{k}(t)-\overline{\theta}(t))(\theta_{k}(t)-\overline{\theta}(t))^{T}$
            \State $y_j(t+1) \longleftarrow y_j(t) + 2 X(t) r_{a(t),j}(t)$, and $\overline{\mu_j(t+1)} = B_j(t+1)^{-1} y_j(t+1)$
            \State Update posterior distribution $p(\delta_n(t+1))$ $p(\alpha_n(t+1))$ $p(\beta_n(t+1))$ $p(\sigma_n(t+1))$ by equations
            \State Update Q function $Q^h (x^h,a^h) = E[r^h|x^h,a^h] + max_{a^{h+1}} Q^{h+1}(x^{h+1},a^{h+1})$
            \EndIf
            \State episode t using greedy algorithm $\pi(\theta(t))$ of $a^h = a^h(\theta(t),j_t^h)$
       \EndFor
       \State Observe trajectory $[j_t,a_t,r_t]$
    \EndFor

    \end{algorithmic}
\end{algorithm}

\subsection{Deep Deterministic Policy Gradient with Contextual Information}


The Thompson sampling algorithm is only suitable for environments with discrete actions and states. In this section, we demonstrate a reinforcement learning algorithm called DDPG that is also based on the AC framework and can adapt to continuous action and state environments. In response to the slow iteration speed of the DDPG algorithm and the inability to achieve fast transactions, we also design CPPI-DDPG algorithm to balance exploration and exploitation. CPPI strategy is used for providing contextual information and accelerate the iteration speed of DDPG. 

To investigate the randomness of the dynamic stock market, we adopt the following tuple $\langle s, a, r, s'\rangle$ to represent the MDP:

\textbf{State} $s=[p, h, b]$: a vector that includes $D$ kinds of stock price $p \in \mathbb{R}_{+}^{D}$, share $h \in \mathbb{Z}_{+}^{D} $, and the remaining balance $b \in \mathbb{R}_{+}^{D}$.

\textbf{Action} $a$: An action set for $K$ agents.

\textbf{Reward} $r(s,a,s')$: The reward for taking action $a$ given state $s$ and the transition to the new state $s'$.

\textbf{Strategy} $\pi(s, a)$: the strategy of an agent. The basic idea of the policy gradient algorithm is to use a parameterized probability distribution $\pi_{\theta}(a|s) = P(a|s;\theta)$ to represent the policy.


Based on the settings above, we specifically designed a novel loss function according to the CPPI strategies. On the one hand, our ultimate target is to maximize the benefits brought by the sum of strategies. On the other hand, we need each agent to consider its own specific situation for ensuring that the increase or decrease of the overall benefits will not have much influence on its own decisions. At the same time, in order to avoid all agents moving towards the same strategy, we need to set the correlation between agents as part of the loss function to achieve the purpose of portfolio selection. The loss function for agent $i$ can be expressed as, 
\begin{equation}\label{critic loss}
     \mathcal{L}
    (\phi_i)= \lambda \mathbb{E}_{s,a,r,s'}\bigg[(Q_{\phi_i}(s,a)-y)^2\bigg] + (1-\lambda) \sum_{i=1,i \leq j}^{K}Corr(a_i,a_j)^2
\end{equation}
where $a_i$ is the action vector showing the positional confidence vector of agent $i$ under the restriction of strategy CPPI, and $\lambda$ is the hyperparameter that controls the equilibrium.

\begin{algorithm}[H]
\caption{DDPG with quantitative trading strategy}\label{alg:1}
\begin{algorithmic}[1]
\State Initialize $Q_{\phi_i}$, $\pi_{\theta_i}$, $Q_{\phi'_i}$, $\pi_{\theta'_i}$.
\While {training not finished}
    \State Initialize initial state $s$ and a random process $\mathcal{N}$ for\newline
    \hspace*{1.25em} action exploration. 
    \For{each episode}
        \State For each agent select action $a_i=\pi_{\theta_i}(s)+\mathcal{N}_t$
        \State Execute joint action $a$ and observe reward $r$ and \newline
        \hspace*{2.75em} next state $s'$.
        \State Store experience $\langle s,a,r,s' \rangle$ in replay buffer $\mathcal{D}$.
        \State Sample a minibatch of $K$ experiences from $\mathcal{D}$.
        \For {each agent}
            \State Adjust $J(Q)$ using CPPI, then update the \newline
    \hspace*{3em} critic $Q_{\phi_i}$ by minimizing Eq.(\ref{critic loss}).
            \State Update the actor $\pi_{\theta_i}$ using Eq.(\ref{actor loss}).
        \EndFor
        \State Update target networks for each agent:
        \begin{equation*}
            \begin{split}
                \phi'_i\leftarrow \tau \phi_i+(1-\tau)\phi'_i,\\
                \theta'_i\leftarrow \tau \theta_i+(1-\tau)\theta'_i.
                \vspace{-.2cm}
            \end{split}
        \end{equation*}
    \EndFor
\EndWhile
\end{algorithmic}
\end{algorithm}

\section{Experiments}\label{sec4}

\subsection{Dataset and Settings}

In our experiment, the operation of algorithm with contextual information is modeled, the object is to minimize the regret bound, with the side information. To demonstrate the impact of contextual information, we compared this algorithm with the original Asymmetric-TS using the dataset from previous experiments with side information. For synthetic asymmetric, we generated contexts $x_{t,a} \in R^{10}$ from alpha-stable distributions for all arms.

When dealing with stock prices, there are different approaches to consider when using side information. These approaches depend on the type of information required to make accurate predictions in the financial domain. At present, the application of RL in quantitative trading in academia can be roughly divided into four types: Portfolio Management, Single asset trading signal, Execution, and Option hedging. Portfolio Management is generally low-frequency trading and Execution is generally based on high-frequency tick level data strategies.

For low-frequency trading data, shares are listed Exchange through Tushare in Shenzhen Stock using Python had been chosen as risk assets. Price to earnings ratio, price to book ratio, price to sales ratio, price to cash ratio, operating revenue growth rate, operating profit growth rate, sales net profit margin, gross profit margin, previous period rise and fall, and circulating market value are selected as contextual information.

For high-frequency tick level data, two datasets with different numbers of stocks included in the S$\&$P 500 index are selected. We obtain stock data that consist of opening, high, low, closing, and volume values from Yahoo Finance and use a smaller dataset to compare with other methods in detail. As output data for training, the first 70$\%$ of the data in this interval is used as training data, and the last 30$\%$ is used as testing data.

Together with the cash as the risk-free asset, the investment products to be managed may exponentially increase. The data in the training set is from January.1st 2018 to December.31st 2020 while the data in the testing set is from January.1st 2021 to December.31st 2021. In order to better fit the actual situation of the market, we have imposed restrictions on the data such as non-negative remaining balance and transaction cost. We initialize our cash and aim to get the highest profits with the trading strategies mentioned above.

When dealing with recommendation data, we choose the MovieLen 100K dataset. The version of the MovieLen 100K dataset includes several features with the “-ratings” suffix, including: “$movie_{id}$”: A unique identifier for the rated movie; “$user_{rating}$”: The score given by the user for the movie on a five-star scale; “$user_{gender}$”: The gender of the user who made the rating, with “true” values corresponding to male “$bucketized_{user_{age}}$”: Buckets of age values of the user who made the rating and so on. Other versions of the MovieLens dataset may include additional features such as movie genres, year of release, and user occupation. These features can be used to create more complex recommendation systems that take contextual information into consideration when making movie recommendations.

\subsection{Experiments for Asymmetric Alpha-Thompson Sampling with Linear Contextual Bandits  }

The main difference between contextual multi-armed bandits algorithm and non-contextual one is that the linear relationship leads to the uncertainty of $\theta$.

The experimental results show that the context information can have a great impact on the regret bound, and help to extract the information faster. At the same time, due to the complex relationship between internal factors, it is more unstable than the model without context information. 

For synthetic data  with contextual information, asymmetric $\alpha$-Thompson algorithm performs much better than symmetric $\alpha$-Thompson sampling with Linear Contextual Bandits as the dataset is generated based on the asymmetric $\alpha$-stable distribution. For stock price with contextual information, the symmetric $\alpha$-Thompson sampling with Linear Contextual Bandits performs better as contextual information is extremely important for stock prices. For recommendation data, the asymmetric $\alpha$-Thompson algorithm performs relatively better than symmetric $\alpha$-Thompson sampling with Linear Contextual Bandits.
\begin{figure}[h]
  \centering
  \includegraphics[width=0.75\linewidth]{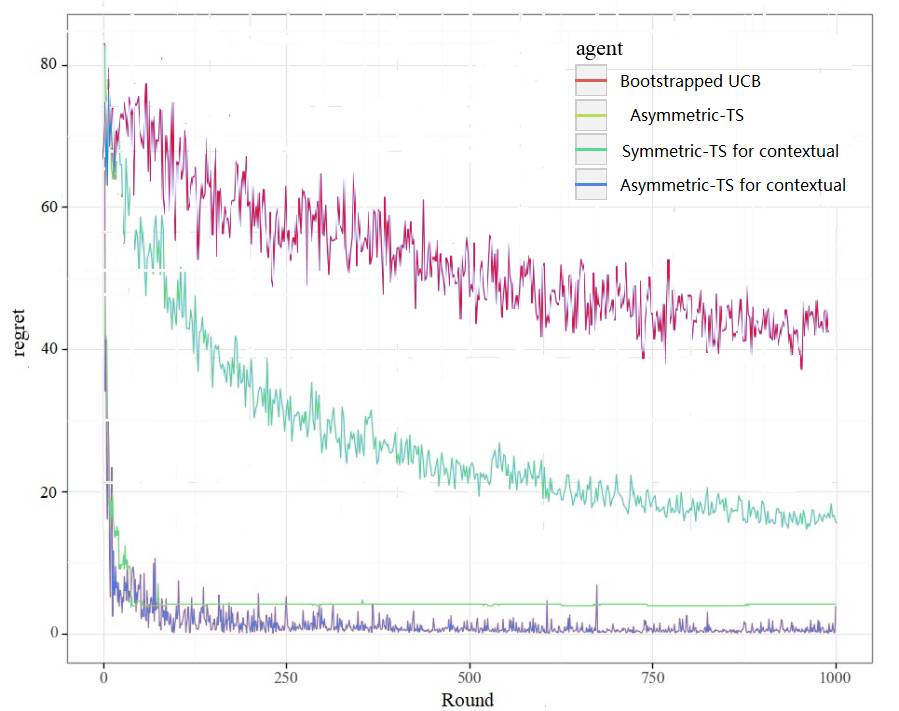}
  \caption[regret for asymmetric data with contextual information]{Regret for asymmetric data with contextual information}
  \label{figure11}
\end{figure}

\begin{figure}[H]
  \centering
  \includegraphics[width=0.75\linewidth]{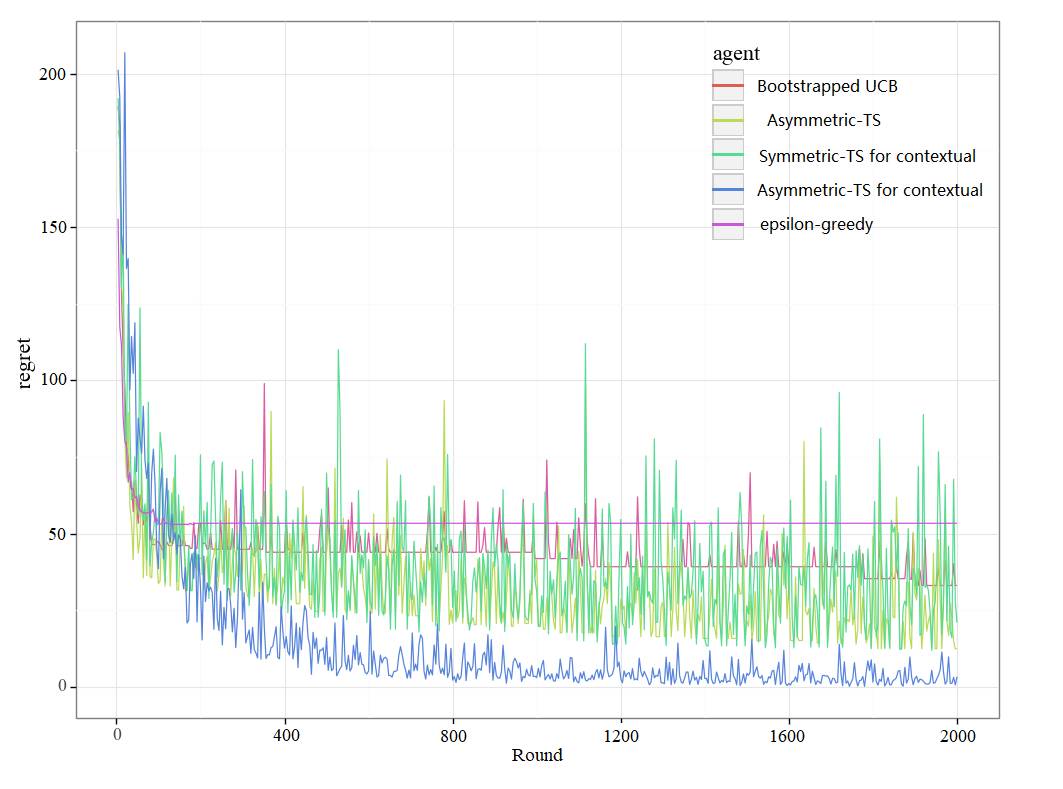}
  \caption[regret for recommendation data with contextual information]{Regret for stock selection with contextual information}
  \label{figure12}
\end{figure}

\begin{figure}[H]
  \centering
  \includegraphics[width=0.75\linewidth]{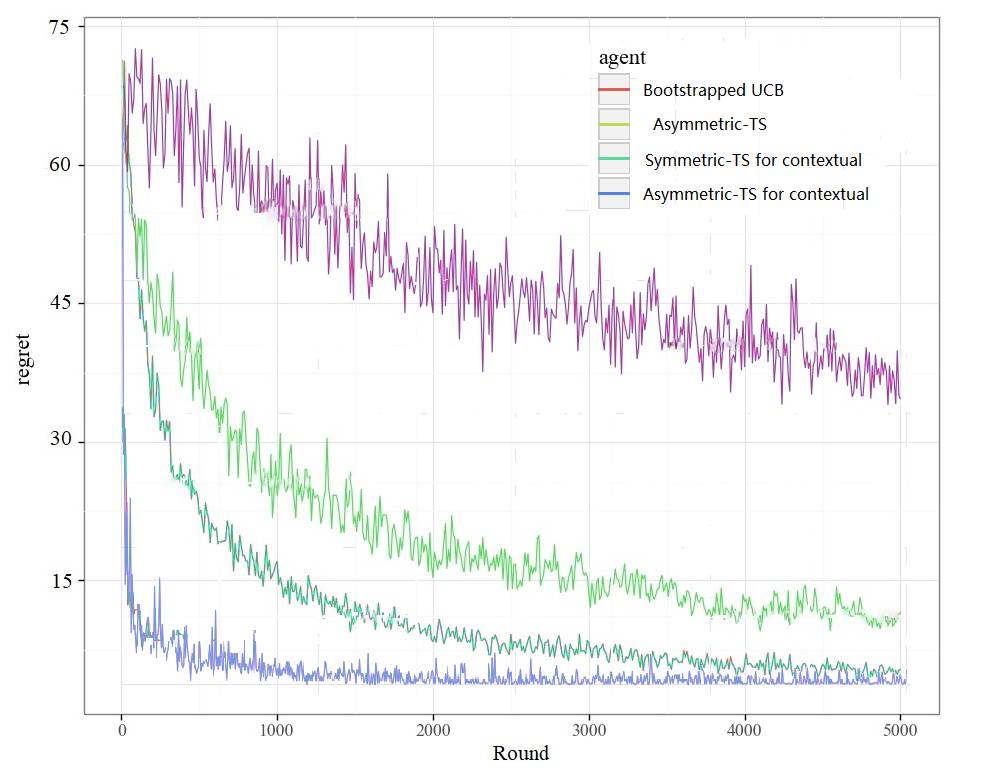}
  \caption[regret for recommendation data with contextual information]{Regret for recommendation data with contextual information}
  \label{figure13}
\end{figure}

\subsection{Experiments for Asymmetric Alpha-Thompson Sampling with Adversarial Contextual Bandits }

Adversarial Contextual Bandits involve the impact of the agent's own actions on future choices and even the environment, which can cause fluctuations in its Regret Bound. In order to reflect the impact of actions on the overall environment, an interference term needs to be added to the existing data. This enables the evaluation of algorithms in the context of Adversary Contextual Bandits, where the algorithm’s decisions not only affect the user receiving the recommendation but also impact the recommendation environment for the next user. The originally optimized action may actually lead to an increase in the Regret Bound. To better illustrate the optimization process, we compare the Regret Bound during each iteration in the table.

The number on the left represents the winning rate of the algorithm on the vertical axis, while the number on the right represents the winning rate of the algorithm on the horizontal axis. The table shows the accuracy of different algorithms compared to other algorithms in dynamic systems. The AC-TS algorithm suitable for MDP, although not performing as well as reinforcement learning algorithms such as DQL and QL, is superior to CB-TS algorithm that does not provide feedback with action changes. 
\begin{table}[h]
    \vspace{-0.5cm}
    \normalsize
    \small
    \caption{Average Wins for different reinforcement learning}
    \label{tab:data}
    \centering
    \begin{tabular}{c c c c c c}
      \toprule
      RL & QL & DQL & SARSA & CB-TS & AC-TS \\
      \midrule 
     QL &  - & 62:38 & 55:45 & 63:37 & 54:46 \\
     DQL  & 38 : 62 & - & 40:60 & 48:52 & 48:52 \\
     SARSA & 45 : 55 & 60:40 & - & 63:37 & 51:49 \\
     CB-TS & 37:63 & 52:48 & 37:63 & - & 42:58 \\
     AC-TS & 46:54 & 52:48 & 49:51 & 58:42 & - \\
    avg wins(\%) & 55.1 & 45.2 & 53.4 & 40.5 & 48 \\
      \bottomrule
    \end{tabular}
\end{table}

\begin{figure}[H]
  \centering
  \includegraphics[width=0.7\linewidth]{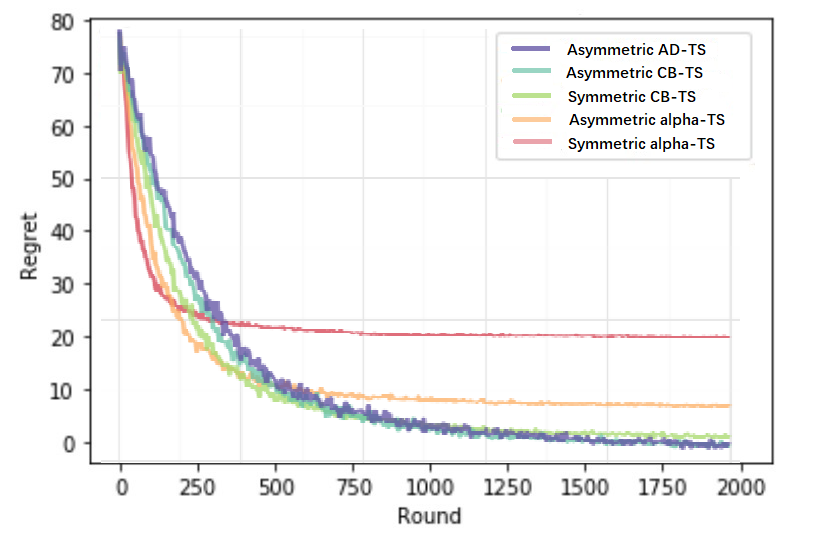}
  \caption{Performance of Adversarial Contextual bandits}
  \label{figure13}
\end{figure}

These figures show that the variation process of the regret bound of each algorithm during the iteration process. Due to the impact of actions on the environment, there are often fluctuations in the early stages of the iteration. The poor performance of CB-TS is because it cannot cover the impact of actions. Sometimes the choice of actions can actually lead to fluctuations in Regret Bound, but in the long run, the impact of actions is a factor that needs to be considered in dynamic systems. The lower the Regret Bound, the closer it is to the theoretical optimal action selection.

\subsection{Portfolio Management}

To study how each agent has made a series of trading decisions over time in the test phase, we visualize the general trading behavior for each agent on 100 shares with DDPG, CPPI-DDPG, and AD-TS, respectively. 
As shown in Fig.~\ref{fig:theo}, the thermodynamic diagram presents how the agents with different strategies choose to allocate the asset. The agents with DDPG prefer the relatively uniform allocation while the assets allocated by those with CPPI-DDPG and AD-TS are more sparse. The sparsity of CPPI-DDPG stems from its strategy itself, as its aversion to risk prevents it from incorporating high-risk stocks into its investment portfolio. The sparsity of AD-TS mainly stems from the discontinuity of its actions and states, which cannot be freely selected like the DDPG algorithm.

\begin{figure}[h]
    \centering
    \includegraphics[width=.8\linewidth]{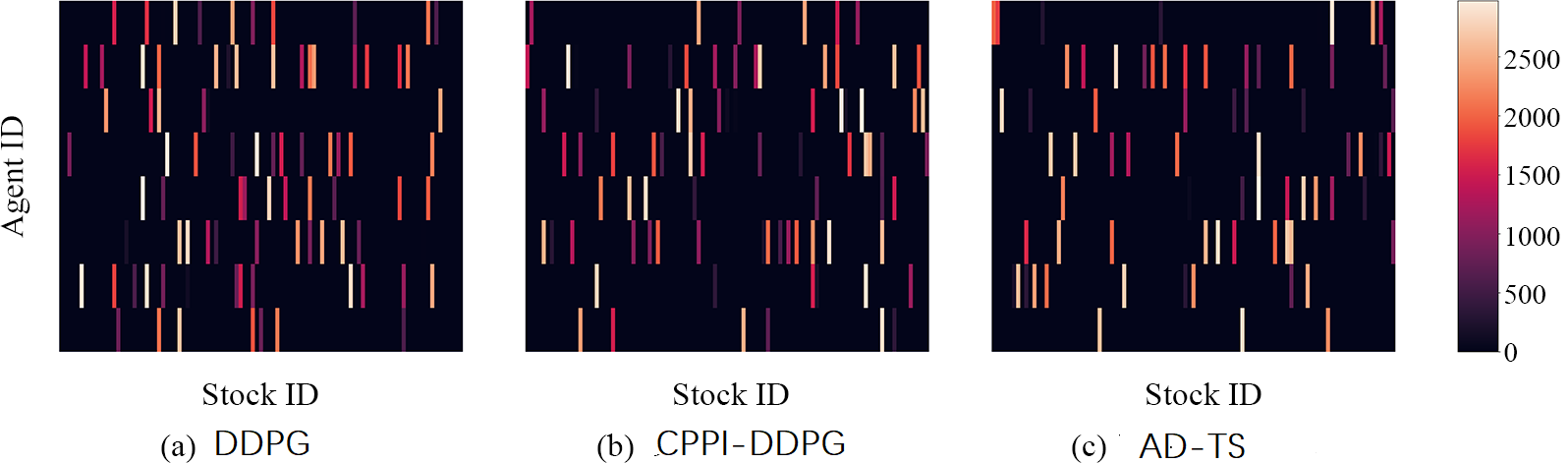}
    \caption{The asset allocations with DDPG, CPPI-DDPG, and AD-TS strategies .}
    \label{fig:theo}
\end{figure}

\begin{figure}[h]
\centering
\includegraphics[width=.8\linewidth]{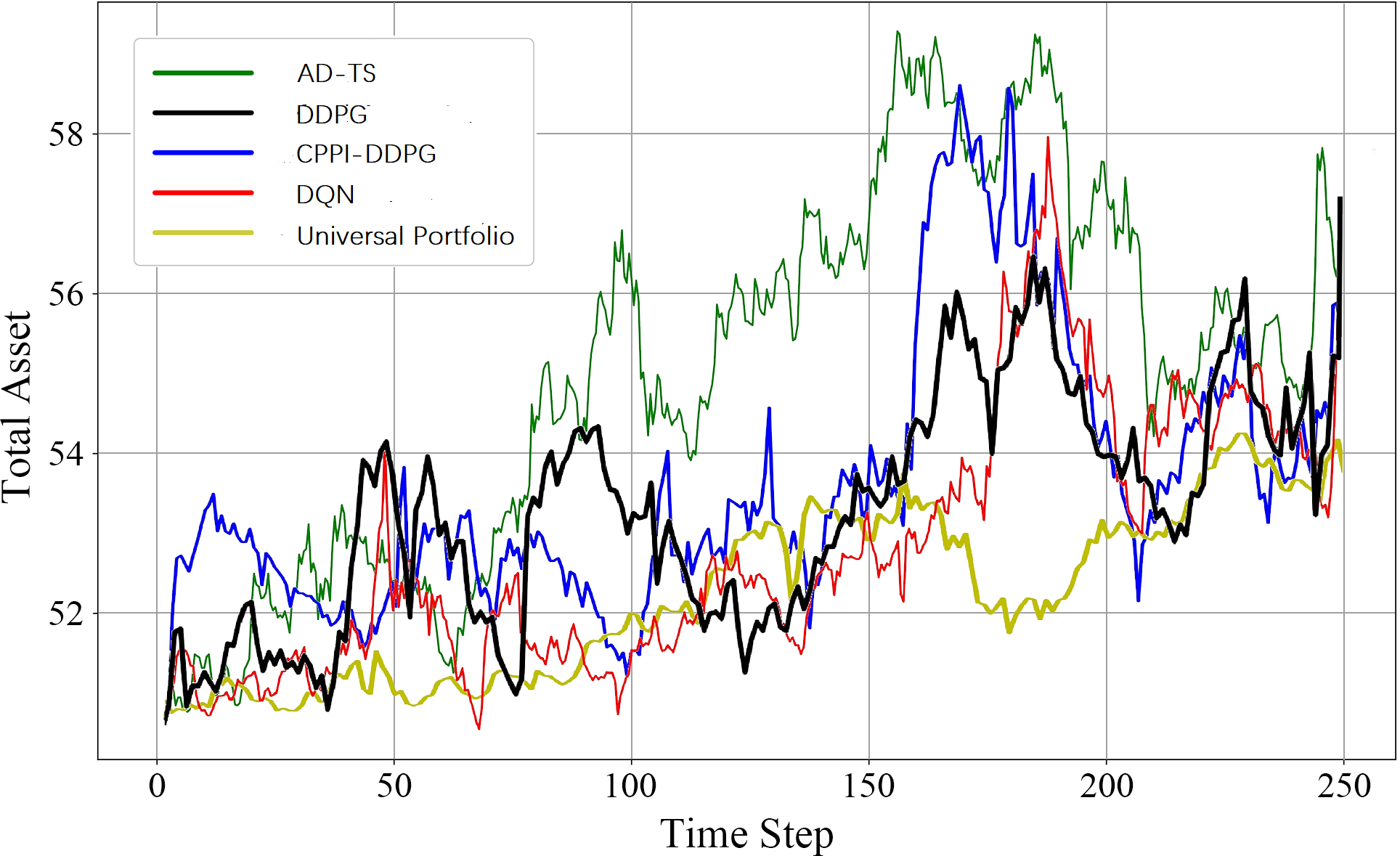}\caption{The short-term performances of portfolios with different strategies (The metrics of Total Asset and Time Step: $10^3$ RMB and Day).}
\label{reward}
\end{figure}

\begin{figure}[h]
\centering
\includegraphics[width=.75\linewidth]{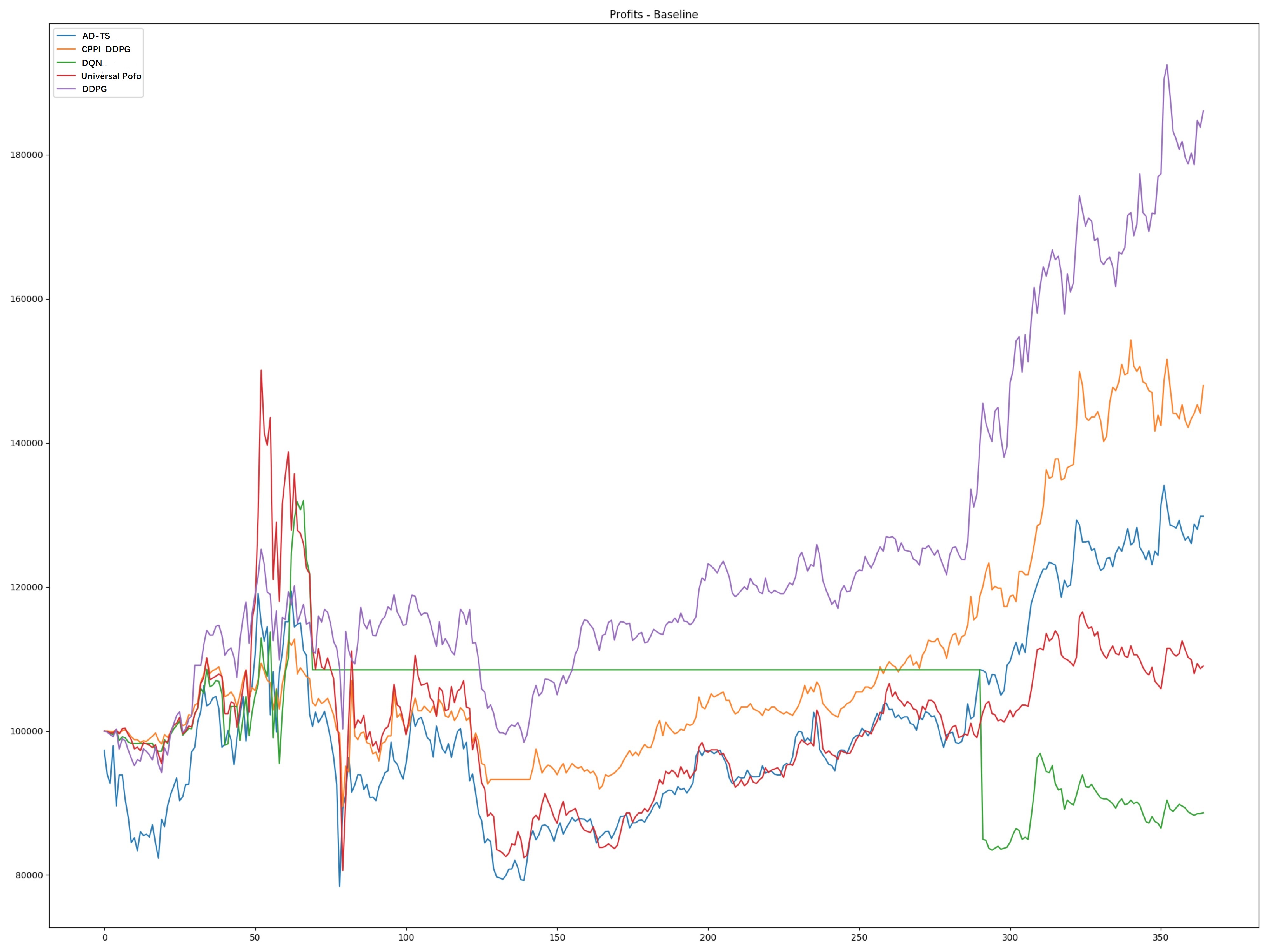}\caption{The long-term performances of portfolios with different strategies.}
\label{reward1}
\end{figure}

The changes of the total assets with different trading strategies over time in short term are present in Fig.~\ref{reward}. In the short term, the advantages of the AD-TS algorithm are more obvious, and based on the assumption of the reward distribution function, its iteration speed is faster than other algorithms. The CPPI-DDPG algorithm, due to its model based nature, outperforms the DDPG algorithm in the short term under supervised conditions.

As for long-term performance shown in Fig.~\ref{reward1}, the advantage of DDPG not being restricted by supervision is reflected, and it has the best performance. Due to a lack of exploration of the external environment, DQN has fallen into suboptimal solutions and has adopted a long-term strategy of not trading.

Finally, we compare the performance of our trading strategies to that of Universal Portfolios (UP), DQN and DDPG through Annual Return (AR), Sharpe Ratio (SR), and Maximum Drawdown (MaxD, namely the maximum portfolio value loss from the peak to the bottom).The performance of AR, SR and MaxD are given in Table~\ref{tab:data}. 

The UP is a common portfolio method, which makes optimal decisions through the calculation of the correlation of different stock returns. However, it cannot cope with real-time data and performs poorly in the test set. The problem of DQN is the lack of exploration ability. There are too many uncertain factors in the stock market for the strategy obtained by single agent. CPPI-DDPG and AD-TS can also degenerate into the classic DDPG strategy under specific parameters, and they can adjust their parameters according to the investors' individual risk preferences.
\begin{table}[ht]
    \normalsize
    \caption{Comparison of Different Strategies}
    \label{tab:data}
    \centering
    \begin{tabular}{c c c c}
      \toprule
      Strategy & AR & SR & MaxD \\
      \midrule 
      UP & $3.36\%$ & $9.2\%$ & $3.48\%$\\
      DQN & $6.47\%$ & $8.3\%$ & $6.35\%$\\
      DDPG & $8.22\%$ & $11.7\%$ & $4.79\%$\\
      CPPI-DDPG & $7.76\%$ & $\mathbf{23.5\%}$ & $\mathbf{3.39\%}$\\
      AD-TS & $\mathbf{9.68\%}$ & $17.8\%$ & $4.5\%$\\
      \bottomrule
    \end{tabular}
\end{table}

\subsection{Execution}

Execution is generally based on high-frequency tick level data strategies. We test algorithmic trading methods using two datasets with different numbers of stocks included in the S$\&$P 500 index. We obtain stock data that consist of opening, high, low, closing, and volume values from Yahoo Finance. 
\begin{figure}[H]
\centering
\includegraphics[width=.7\linewidth]{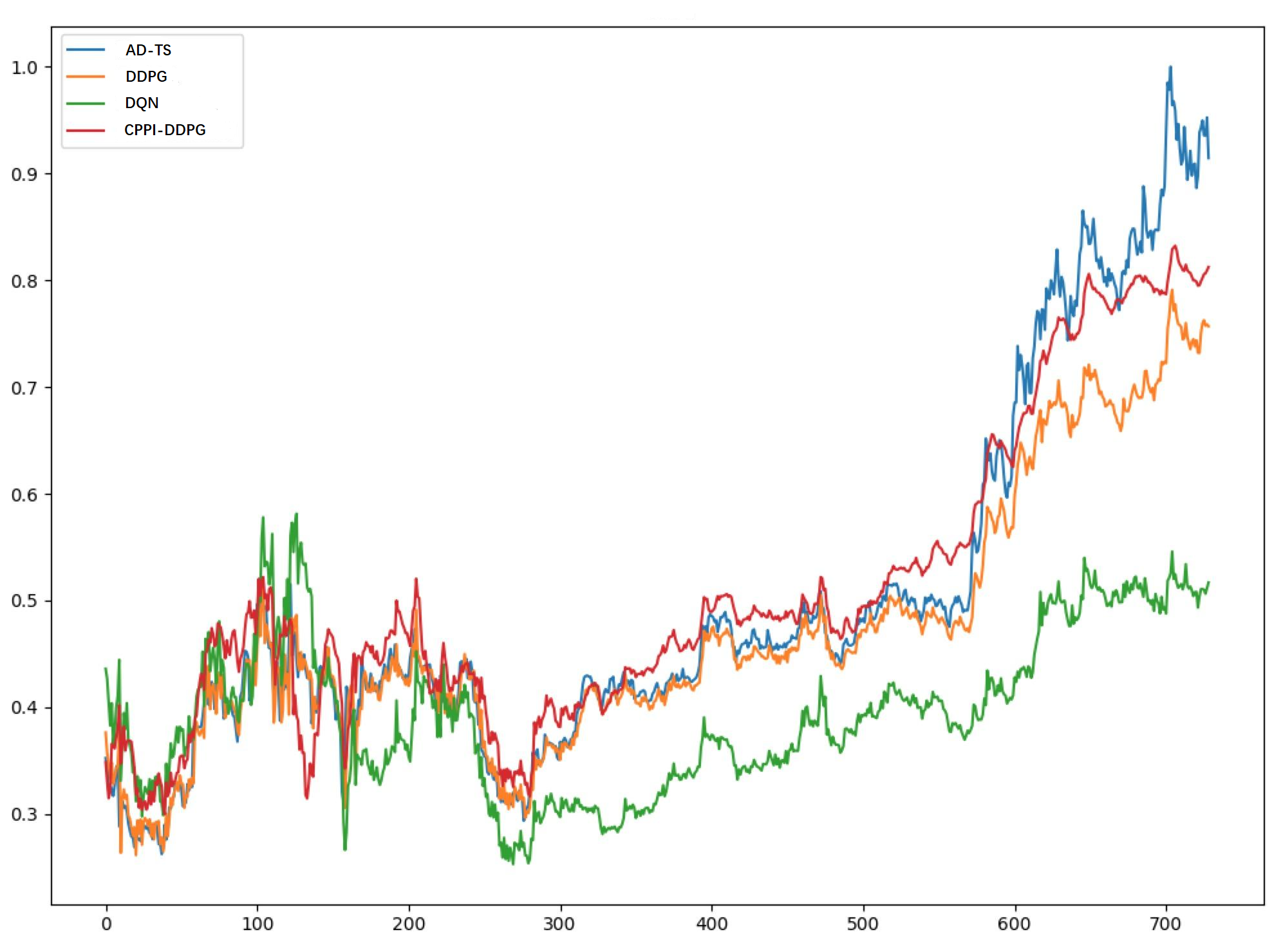}\caption{The short-term performances of executions with different strategies.}
\label{reward2}
\end{figure}

It can be observed that during the Execution, the trading frequency of stocks shifts from low frequency to high frequency, and the impact of the combination between different stocks is reduced. From the short-term and long-term performance, the efficiency and final results of reinforcement learning such as DDPG are not as good as our algorithms. 

Due to its continuous state and continuous actions, DDPG has an advantage in investment portfolios. In high-frequency trading, reinforcement learning such as DDPG still requires supervision (such as CPPI) to improve its efficiency.

\begin{figure}[H]
\centering
\includegraphics[width=.7\linewidth]{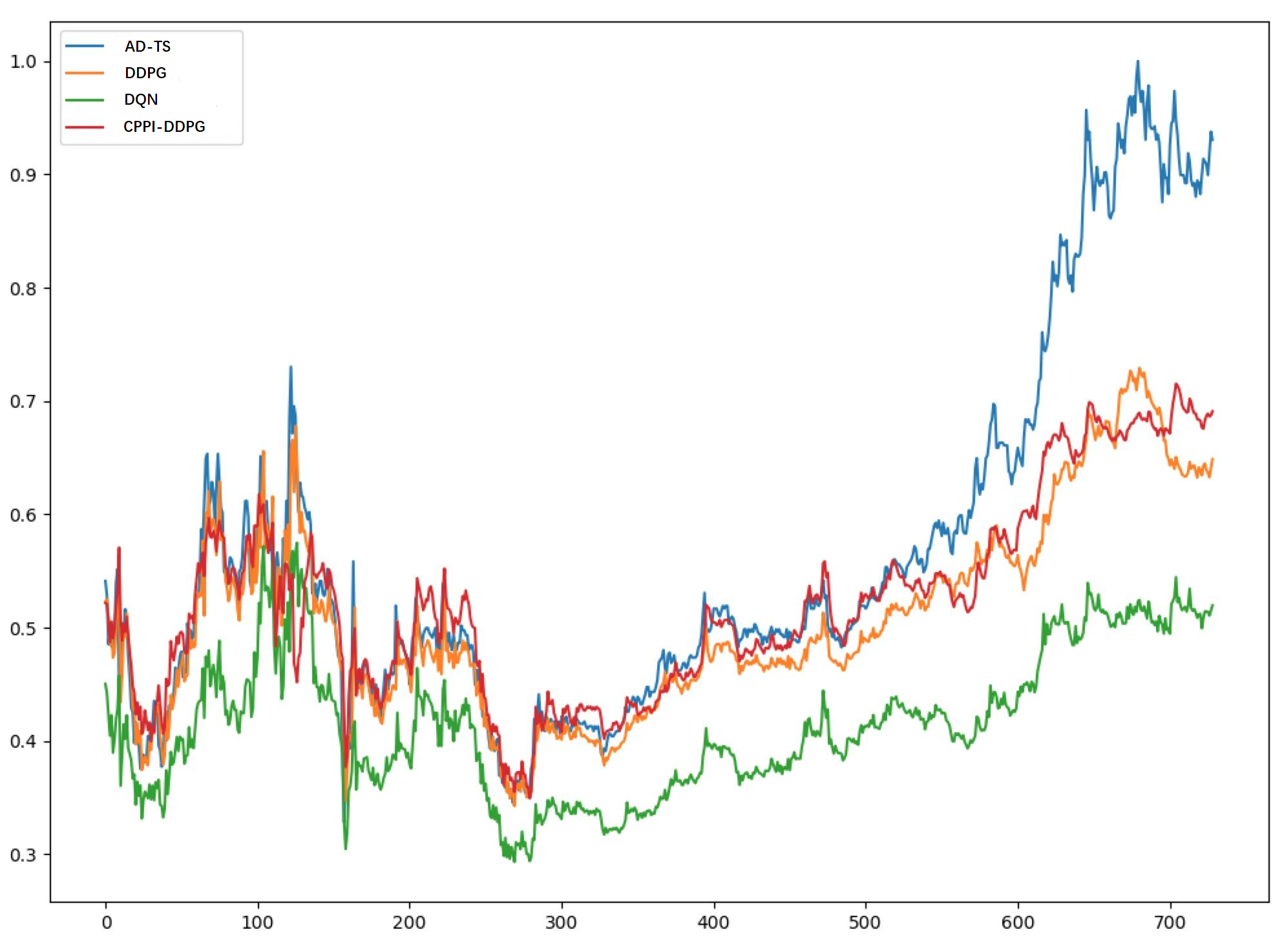}\caption{The long-term performances of executions with different strategies.}
\label{reward3}
\end{figure}

\section{Conclusion}\label{sec5}

In this article, we introduced the derivation of algorithms from MAB to DDPG. MAB, contextual bandit, Q-learning, and DDPG. In actual production, we need to first think about our own assumptions about the problem, and then choose a suitable model based on this assumption. 

For issues in the financial sector, compared with the common gambling machine algorithms such as epsilon greedy, Thompson sampling does not need to manually adjust hyper-parameter, and can better adapt to different problems and reward distribution. Compared to Thompson sampling, reinforcement learning algorithms require more iterations to achieve the same benefits in multi-arm bandits problems. Because Thompson sampling can converge faster through model updates under the Bayesian framework. In addition, reinforcement learning algorithms may lead to a decrease in short-term benefits due to excessive exploration, while Thompson sampling controls the balance between exploration and utilization through the uncertainty of Bayesian models, which can better balance short-term and long-term benefits.

By applying certain supervisory conditions to reinforcement learning, we can avoid excessive exploration in the early stages of learning, thereby improving exploration efficiency. In this article, we chose the CPPI strategy to ensure the efficiency of reinforcement learning in the short term, but it is significantly worse in the long term. In practical operation, we need to understand the specific situation and then choose or adjust the algorithm based on the conditions.

In the future of the work, we would like to adopt a Bayesian framework for the sequential learning \cite{costagli07} and model improvement as agents face more data and decisions. 

\section{Declarations}

\subsection*{Ethical Approval}

This paper belongs to the research of basic algorithms and can not be applied to human or animal studies, there are no ethical issues.
 
\subsection*{Competing interests} 

All authors disclosed no relevant relationships.
 
\subsection*{Authors' contributions} 

Shi Zhendong and Ercan Kuruoğlu wrote the main manuscript text, Xiaoli Wei wrote the theorems part. All authors reviewed the manuscript and participated in the research on the algorithm to be expanded in this manuscript, asymmetric alpha Thompson sampling.
 
\subsection*{Funding} 

This study did not receive support from funding.
 
\subsection*{Availability of data and materials}

We have provided specific sources for the all data generated or analysed during this study. The synthesized data was generated using Python through Chamber's research. The social data that support the findings of this study are openly available in the tushare package in Python and MovieLen 100K dataset. 

\bibliography{sn-bibliography}

\begin{thebibliography}{32}
\providecommand{\natexlab}[1]{#1}
\providecommand{\url}[1]{{#1}}
\providecommand{\urlprefix}{URL }
\providecommand{\doi}[1]{\url{https://doi.org/#1}}
\providecommand{\eprint}[2][]{\url{#2}}
 \bibcommenthead

\bibitem[{Agrawal and Goyal(2013)}]{bib1}
Agrawal S, Goyal N (2013) Thompson sampling for contextual bandits with linear
  payoffs. International conference on machine learning pp 127--135

\bibitem[{An et~al(2022)An, Sun, and Wang}]{bib2}
An B, Sun S, Wang R (2022) Deep reinforcement learning for quantitative
  trading: Challenges and opportunities. IEEE Intelligent Systems 37(2):23--26

\bibitem[{Auer et~al(1995)Auer, Cesa-Bianchi, Freund, and Schapire}]{bib3}
Auer P, Cesa-Bianchi N, Freund Y, et~al (1995) Gambling in a rigged casino: The
  adversarial multi-armed bandit problem. Proceedings of IEEE 36th annual
  foundations of computer science pp 322--331

\bibitem[{Baisero and Amato(2021)}]{baisero2021unbiased}
Baisero A, Amato C (2021) Unbiased {A}symmetric {A}ctor-{C}ritic for
  {P}artially {O}bservable {R}einforcement {L}earning. The Computing Research
  Repository

\bibitem[{Balder et~al(2009)Balder, Brandl, and Mahayni}]{bib4}
Balder S, Brandl M, Mahayni A (2009) Effectiveness of cppi strategies under
  discrete-time trading. Journal of Economic Dynamics and Control
  33(1):204--220

\bibitem[{Bubeck et~al(2013)Bubeck, Cesa-Bianchi, and Lugosi}]{bib27}
Bubeck S, Cesa-Bianchi N, Lugosi G (2013) Bandits with heavy tail. IEEE
  Transactions on Information Theory 59(11):7711--7717

\bibitem[{Cappé et~al(2013)Cappé, Garivier, Maillard, Munos, and
  Stoltz}]{bib5}
Cappé O, Garivier A, Maillard OA, et~al (2013) {Kullback-Leibler upper
  confidence bounds for optimal sequential allocation}. The Annals of
  Statistics p 1516–1541

\bibitem[{Chen et~al(2016)Chen, So, and Kuruoglu}]{chen2016variance}
Chen Y, So HC, Kuruoglu EE (2016) Variance analysis of unbiased least lp-norm
  estimator in non-gaussian noise. Signal Processing 122:190--203

\bibitem[{Chu et~al(2011)Chu, Li, Reyzin, and Schapire}]{bib6}
Chu W, Li L, Reyzin L, et~al (2011) Contextual bandits with linear payoff
  functions. Proceedings of the Fourteenth International Conference on
  Artificial Intelligence and Statistics pp 208--214

\bibitem[{Costagli and Kuruoğlu(2007)}]{costagli07}
Costagli M, Kuruoğlu EE (2007) Image separation using particle filters.
  Digital Signal Processing 17(5):935--946.
  \doi{https://doi.org/10.1016/j.dsp.2007.04.003},
  \urlprefix\url{https://www.sciencedirect.com/science/article/pii/S1051200407000590},
  special Issue on Bayesian Source Separation

\bibitem[{Daniels et~al(2003)Daniels, Farmer, Gillemot, Iori, and Smith}]{bib7}
Daniels MG, Farmer JD, Gillemot L, et~al (2003) Quantitative model of price
  diffusion and market friction based on trading as a mechanistic random
  process. Physical Review Letters 90(10):108--102

\bibitem[{Dubey and Pentland(2019)}]{bib25}
Dubey A, Pentland A (2019) Thompson {S}ampling on {S}ymmetric alpha-{S}table
  {B}andits. International Joint Conference on Artificial Intelligence

\bibitem[{Fleming and Pang(2004)}]{bib8}
Fleming WH, Pang T (2004) An application of stochastic control theory to
  financial economics. SIAM Journal on Control and Optimization 43(2):502--531

\bibitem[{Gopalan et~al(2014)Gopalan, Mannor, and Mansour}]{bib9}
Gopalan A, Mannor S, Mansour Y (2014) Thompson sampling for complex online
  problems. International conference on machine learning pp 100--108

\bibitem[{Guo et~al(2017)Guo, Lai, Shek, and Wong}]{bib10}
Guo X, Lai TL, Shek H, et~al (2017) Quantitative trading: Algorithms,
  Analytics, Data, Models, Optimization. CRC Press

\bibitem[{Gupta(2014)}]{gupta2014dynamic}
Gupta A (2014) Dynamic sequential decision problems with asymmetric
  information: Some existence results. University of Illinois at
  Urbana-Champaign

\bibitem[{Korda et~al(2013)Korda, Kaufmann, and Munos}]{bib23}
Korda N, Kaufmann E, Munos R (2013) Thompson sampling for 1-dimensional
  exponential family bandits. Advances in neural information processing systems
  26

\bibitem[{Korte and Lov{\'a}sz(1984)}]{bib11}
Korte B, Lov{\'a}sz L (1984) Greedoids-a structural framework for the greedy
  algorithm. Progress in combinatorial optimization pp 221--243

\bibitem[{Kuruoglu(2001)}]{bib26}
Kuruoglu EE (2001) Density parameter estimation of skewed/spl alpha/-stable
  distributions. IEEE Transactions on signal processing 49(10):2192--2201

\bibitem[{Kuruoglu(2003)}]{kuruoglu03}
Kuruoglu EE (2003) Analytical representation for positive /spl alpha/-stable
  densities. In: 2003 IEEE International Conference on Acoustics, Speech, and
  Signal Processing, 2003. Proceedings. (ICASSP '03)., pp VI--729

\bibitem[{Li et~al(2022)Li, Cui, Cao, Du, and Zhang}]{bib12}
Li X, Cui C, Cao D, et~al (2022) Hypergraph-based reinforcement learning for
  stock portfolio selection. 2022 IEEE International Conference on Acoustics,
  Speech and Signal Processing (ICASSP) pp 4028--4032

\bibitem[{Mnih et~al(2015)Mnih, Kavukcuoglu, Silver, Rusu, Veness, Bellemare,
  Graves, Riedmiller, Fidjeland, Ostrovski et~al}]{bib14}
Mnih V, Kavukcuoglu K, Silver D, et~al (2015) Human-level control through deep
  reinforcement learning. Nature 518(7540):529--533

\bibitem[{Moskowitz et~al(2012)Moskowitz, Ooi, and Pedersen}]{bib15}
Moskowitz TJ, Ooi YH, Pedersen LH (2012) Time series momentum. Journal of
  financial economics 104(2):228--250

\bibitem[{Pham(2009)}]{bib16}
Pham H (2009) Continuous-time Stochastic Control and Optimization with
  Financial Applications, vol~61. Springer Science \& Business Media

\bibitem[{Russo and Van(2014{\natexlab{a}})}]{bib17}
Russo D, Van R (2014{\natexlab{a}}) Learning to optimize via posterior
  sampling. Mathematics of Operations Research 39(4):1221--1243

\bibitem[{Russo and Van(2014{\natexlab{b}})}]{bib22}
Russo D, Van R (2014{\natexlab{b}}) Learning to optimize via posterior
  sampling. Mathematics of Operations Research 39(4):1221--1243

\bibitem[{Samorodnitsky and Taqqu(1997)}]{samorodnitsky1997stable}
Samorodnitsky G, Taqqu M (1997) Stable non-gauss/an random processes.
  Econometric Theory 13:133--142

\bibitem[{Shi et~al(2022)Shi, Kuruoglu, and Wei}]{bib24}
Shi Z, Kuruoglu E, Wei X (2022) Thompson sampling on asymmetric $\alpha$-stable
  bandits. In Proceedings of the 15th International Conference on Agents and
  Artificial Intelligence 3:434--441

\bibitem[{Sutton and Barto(2018)}]{bib18}
Sutton RS, Barto AG (2018) Reinforcement learning: An introduction. MIT press

\bibitem[{Thakkar and Chaudhari(2021)}]{bib19}
Thakkar A, Chaudhari K (2021) A comprehensive survey on deep neural networks
  for stock market: The need, challenges, and future directions. Expert Systems
  with Applications 177:114800

\bibitem[{Tsantekidis et~al(2021)Tsantekidis, Passalis, and Tefas}]{bib20}
Tsantekidis A, Passalis N, Tefas A (2021) Diversity-driven knowledge
  distillation for financial trading using deep reinforcement learning. Neural
  Networks 140:193--202

\bibitem[{Zhang et~al(2021)Zhang, Jiang, and Su}]{bib21}
Zhang H, Jiang Z, Su J (2021) A deep deterministic policy gradient-based
  strategy for stocks portfolio management. 2021 IEEE 6th International
  Conference on Big Data Analytics (ICBDA) pp 230--238

\end{thebibliography}

\end{document}